\documentclass[a4paper,fleqn]{cas-dc}
\usepackage{xcolor}%定义了一些颜色
\usepackage{colortbl,booktabs}%第二个包定义了几个*rule
\usepackage[authoryear,longnamesfirst]{natbib}
\usepackage{graphicx}
\usepackage{epstopdf}       % eps图片文件读取
\usepackage{amsmath}
\def\tsc#1{\csdef{#1}{\textsc{\lowercase{#1}}\xspace}}
\tsc{WGM}
\tsc{QE}
\tsc{EP}
\tsc{PMS}
\tsc{BEC}
\tsc{DE}
\usepackage{hyperref}
\hypersetup{colorlinks,citecolor=blue,linkcolor=blue,urlcolor=Blue}
\usepackage{caption}
\usepackage{float}
\usepackage{fancyhdr}
\begin{document}
% 下面一行代码作用是保证每行文字都自动换行
\begin{sloppypar}
\let\WriteBookmarks\relax
\def\floatpagepagefraction{1}
\def\textpagefraction{.001}
% \bibliographystyle{plain} 
% \bibliography{cas-refs} 

% 页眉小标题
\shorttitle{}

%\shortauthors{CV Radhakrishnan et~al.}

\title [mode = title]{ \centering CAINNFlow: Convolutional block Attention modules and Invertible Neural Networks Flow for anomaly detection and localization tasks} 

%下标脚注
%\tnotemark[1,2]


%下标脚注解释
% \tnotetext[1]{e-mail: yrqUni@gmail.com}

% \tnotetext[2]{The second title footnote which is a longer text matter
%   to fill through the whole text width and overflow into
%   another line in the footnotes area of the first page.}

% \author[1,3]{Author Name}[type=editor,
% \author[1,3]{CV Radhakrishnan}[type=editor,
%                         auid=000,bioid=1,
%                         prefix=Sir,
%                         role=Researcher,
%                         orcid=0000-0001-7511-2910]

% \author[1]{Ruiqing Yan}

% \author{Ruiqing Yan \authorrefmark{1}}

\address[1]{BIPT,Qingyuan North Road 19,BeiJing,102627,China}
\address[2]{Computer Network Information Center, Chinese Academy of Sciences,100190,China}
\address[3]{University of Chinese Academy of Sciences,BeiJing,100049,China}
\address[4]{Ant Group,China}

% \corresp{Corresponding author: Fang Wang (e-mail: fangwang@bipt.edu.cn)}
\cortext[0]{E-mail: \href{https://yrqUni@gmail.com}{ yrqUni@gmail.com} (R. Yan)}

\author[1]{Ruiqing Yan}[type=editor,
                        auid=000,
                        orcid=0000-0003-2798-4655,
                        % e-mail=yrqUni@gmail.com,
                        % linkedin=yrqUni@gmail.com,
                        ]
\fnmark[1]

% \ead[url]{  https://orcid.org/0000-0003-2798-4655}                       
% \ead[url]{yrqUni@gmail.com}
\author[1]{Fan Zhang}[style=chinese]
\fnmark[2]
\author[1]{Mengyuan Huang}[
%   suffix=Jr,
  ]
\fnmark[3]
\author[1]{Wu Liu}[
%   suffix=Jr,
  ]
\fnmark[3]
\author[1]{Dongyu Hu}[
%   suffix=Jr,
  ]
\fnmark[3]
\author[1]{Jinfeng Li}[
%   suffix=Jr,
  ]
\fnmark[3]
\author[1]{Qiang Liu}[
%   suffix=Jr,
  ]
\fnmark[4]
\author[2,3]{Jinrong Jiang}[
%   suffix=Jr,
  ]
\cormark[1]
\fnmark[4]
\author[1]{Qianjin Guo}[
    % orcid=0000-0002-8895-7899,
%   suffix=Jr,
  ]
%   \ead[URL]{guoqianjin@bipt.edu.cn}
%   \ead[URL]{https://orcid.org/0000-0002-8895-7899}   
\fnmark[4]
\cormark[1]
\author[4]{Linghan Zheng}[
%   suffix=Jr,
  ]
\fnmark[4]
% ++++++++++++++++++++++++++++++++++++++++++++++++++++++++++
% 作者标注
% \ead{cvr3@sayahna.org}
% \ead[URL]{www.sayahna.org}

% \credit{Data curation, Writing - Original draft preparation}

% \affiliation[2]{organization={Sayahna Foundation},
%     % addressline={}, 
%     city={Jagathy},
%     % citysep={}, % Uncomment if no comma needed between city and postcode
%     postcode={695014}, 
%     state={Trivandrum},
%     country={India}}

% \author%
% [1,3]
% {Rishi T.}
% \cormark[2]
% \fnmark[1,3]
% \ead{rishi@stmdocs.in}
% \ead[URL]{www.stmdocs.in}

% \affiliation[3]{organization={STM Document Engineering Pvt Ltd.},
%     addressline={Mepukada}, 
%     city={Malayinkil},
%     % citysep={}, % Uncomment if no comma needed between city and postcode
%     postcode={695571}, 
%     state={Trivandrum},
%     country={India}}

%通讯作者
% \cortext[cor1]{Corresponding author}
% \cortext[cor2]{Principal corresponding author}

% % Footnote text
% \fntext[fn1]{This is the first author footnote. but is common to third
% \fntext[fn2]{Another author footnote, this is a very long footnote and
%   it should be a really long footnote. But this footnote is not yet
%   sufficiently long enough to make two lines of footnote text.}

% \nonumnote{This note has no numbers. In this work we demonstrate $a_b$
%   the formation Y\_1 of a new type of polariton on the interface
%   between a cuprous oxide slab and a polystyrene micro-sphere placed
\maketitle 

% 指定页的页眉页脚设置
\thispagestyle{fancy}
\rhead{\includegraphics[scale=0.02]{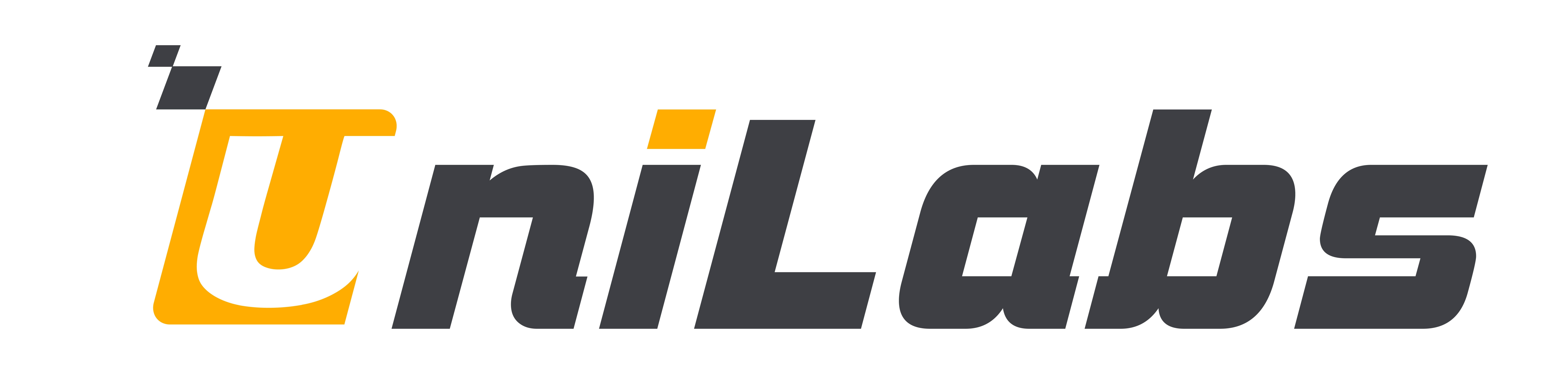}}
\lhead{}
\lfoot{}
% <span style="font-family: Arial, Helvetica, sans-serif;"><span style="color:#ff000;">thispagestyle{fancy}</span></span>
\begin{abstract}
Detection of object anomalies is crucial in industrial processes, but unsupervised anomaly detection and localization is particularly important due to the difficulty of obtaining a large number of defective samples and the unpredictable types of anomalies in real life. Among the existing unsupervised anomaly detection and localization methods, the NF-based scheme has achieved better results. However, the two subnets (complex functions) $s_{i}(u_{i})$ and $t_{i}(u_{i})$ in NF are usually multilayer perceptrons, which need to squeeze the input visual features from 2D flattening to 1D, destroying the spatial location relationship in the feature map and losing the spatial structure information. 
In order to retain and effectively extract spatial structure information, we design in this study a complex function model with alternating CBAM embedded in a stacked $3×3$ full convolution, which is able to retain and effectively extract spatial structure information in the normalized flow model. Extensive experimental results on the MVTec AD dataset show that CAINNFlow achieves advanced levels of accuracy and inference efficiency based on CNN and Transformer backbone networks as feature extractors, and CAINNFlow achieves a pixel-level AUC of $98.64\%$ for anomaly detection in MVTec AD.
% In industrial defect detection scenarios, anomalous detection and localization of faulty sections of items are critical. We discovered that most prior studies disregard the link between local and global characteristics while extracting picture features and computing anomaly ratios, which is critical in the anomaly detection process. As a result, we propose the CAINNFlow model structure, which can be used as a plug-in module with arbitrary deep feature extractors, and learn to convert the input visual features into tractable distributions for training through CAINNFlow, and obtain the probability of detecting anomalies in the model estimation stage. CAINNFlow exceeds prior state-of-the-art approaches in terms of accuracy and inference efficiency for diverse backbone networks, according to extensive experimental results on the MVTec AD dataset. In the current study, the model produces the greatest results in anomaly identification.

% This template helps you to create a properly formatted \LaTeX\ manuscript.

% \noindent\texttt{\textbackslash begin{abstract}} \dots 
% \texttt{\textbackslash end{abstract}} and
% \verb+\begin{keyword}+ \verb+...+ \verb+\end{keyword}+ 
% which
% contain the abstract and keywords respectively. 

% \noindent Each keyword shall be separated by a \verb+\sep+ command.
\end{abstract}

% \includegraphics{figs/grabs.pdf}

% \begin{highlights}
% \item Research highlights item 1
% \item Research highlights item 2
% \item Research highlights item 3
% \end{highlights}

% Keywords

\section{Introduction}
Artificial intelligence anomaly detection system is widely used in manufacturing error detection \cite{roth2021towards}, network intrusion detection \cite{bergman2020classification}, and diagnosis of medical cases \cite{wang2021glancing}.Anomaly detection and positioning technology are used to detect and locate anomalies from samples. In the field of industrial manufacturing, it is used to timely find defects in industrial equipment and industrial products that are not easily noticed by people, to ensure product quality and improve industrial production efficiency. In traditional anomaly detection methods, the training sample is roughly divided into samples of normal and abnormal samples, however, in real life, it is difficult to get a lot of defective samples \cite{li2021cutpaste}, the abnormal sample distribution imbalance caused by abnormal types at the same time uncertainly, training can't contain all the abnormal types, so to break through the limitations, we use the form of distribution fitting to train the model with unlabeled data. We use a distribution transformation structure based on a reversible neural network to transform the distribution corresponding to the normal sample into the normal distribution. On the contrary, if the final distribution is not normal, then the input image is abnormal. Our method is mainly composed of a feature extraction module and distribution estimation module, which are responsible for image feature extraction and feature distribution transformation respectively.

In the feature extraction module, we choose spatial structure information of the image, in the field of computer vision, used in spatial structure information of image extraction model mainly divided into two kinds, and based on the Transformer based on convolution, among them, the classic model based on convolution is the ResNet, it through the residual structure can solve the problem of gradient disappeared, it can effectively extract spatial structure information of images. Recently, Transformer based machine vision model has gradually become a new paradigm, leading the traditional model based on Convolutional Neural Network (CNN) in a variety of scenarios. For example, ViT \cite{dosovitskiy2020image} imitates Bert's way of processing natural language and extracts the features of patch embedding sequence by Decoder. Although Transformer based machine vision model has achieved great success, convolution-based ConvNeXt still has faster processing speed and higher accuracy than Swin Transformer \cite{liu2021Swin} under the same FLOPs, which shows us that the CNN structure still has a certain potential. CAINNFlow is a plug-in structure that can be connected to any image feature that extractor for distribution transformation, regardless of the structure based on which image feature extractor is formed.

In the distribution estimation module, we transform the distribution and convert the original probability distribution of the output features of the feature extractor into normal distribution and other distributions (or the original probability distribution itself).In this module, we make assumptions about the distribution of data that based on statistical methods and find out the "anomalies" defined under the assumptions. Recently, a method named NF has been proposed \cite{cunningham2020normalizing}. Flow-based NF can transform an arbitrary probability distribution into another arbitrary probability distribution, Such as the normal distribution converts into other arbitrary distributions, including the normal distribution itself. Improved GAN, VAE, and other traditional methods can not accurately evaluate the probability distribution and reasoning defects. However, the original one-dimensional normalized flow model required flattening the two-dimensional input features into one-dimensional vectors to estimate the distribution, which limited the flow model's ability and destroyed the inherent spatial position relationship of two-dimensional images \cite{cunningham2020normalizing}.To solve this problem, CAINNFlow is proposed, that is, CNN structure embedded in CBAM(Convolutional Block Attention Model) \cite{woo2018cbam} is used to retain the spatial location of input samples, which enables CAINNFlow to effectively extract spatial structure information while preserving spatial structure information. In addition, in the traditional structure of the Convolutional structure, the Attention of each point and channel in the space is the same, which enables the Model to focus on important features and suppress unnecessary features. Using the structure of CBAM, effectively improve the effect of anomaly detection.

\begin{figure*}[]
	\centering
	\includegraphics[width=.95\linewidth]{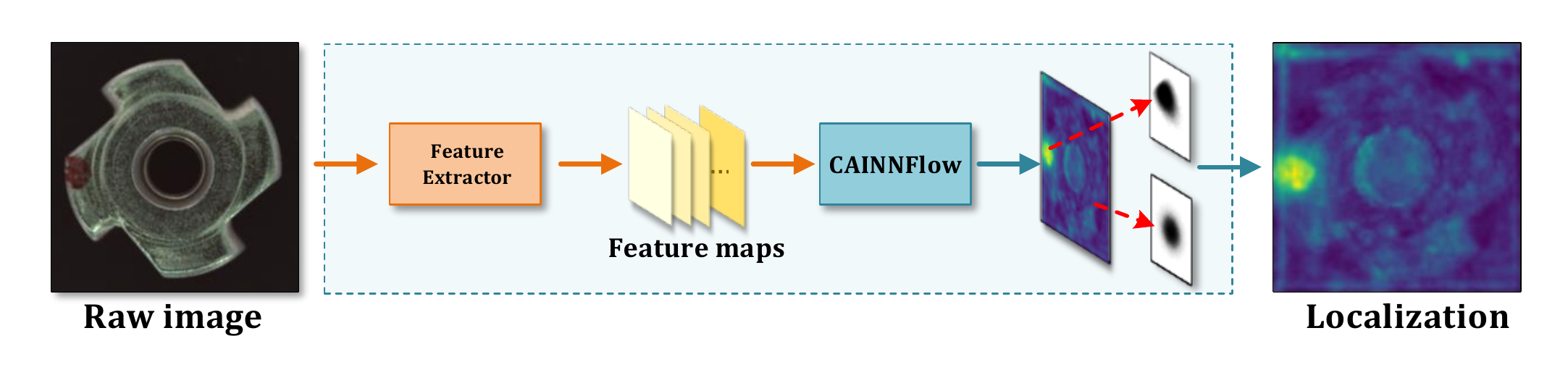}
% 	\includegraphics[width=.95\linewidth]{pic/1.eps}
% 	\includegraphics[height=2.0cm,width=8.5cm]{pic/1.eps}
	%\fbox{\rule[-.5cm]{4cm}{4cm} \rule[-.5cm]{4cm}{0cm}}
	\caption{ An example of the proposed Flow. Flow transforms features of the input image from the original distribution to the standard normal distribution. The features of the normal area in the input image fall in the center of the distribution, while the abnormal features are far away from the center of the distribution}
	\label{fig1}
\end{figure*}

As shown in the figure: \ref{fig1}, we first used the feature extractor pre-trained in the large-scale open field to extract the spatial features of the image, and then input the obtained spatial feature tensor into CAINNFlow to realize anomaly detection and location. During training, CAINNFlow can transform the distribution corresponding to the spatial features of normal images into normal distribution through parameter updating. In the transformation, CAINNFlow can retain the spatial structure information in the features through the stacked CNN structure embedded with CBAM and extract it effectively. During the test, we calculate the distance between the model input image distribution and the standard normal distribution to obtain the anomaly score of each pixel, to achieve pixel-level anomaly detection and location. In general, we use a visual feature extractor based on a neural network for visual feature extraction, and then use CAINNFlow for distribution transformation of reserved spatial information to realize anomaly detection and pixel-level location.

%图片1%%%%%%%%%%%%%%%++++++++++++++++++++
% figure 1
The main contributions of this paper are summarized as follows:

$\bullet$ The CAINNFlow proposed by us retains spatial information during distributed transformation not only uses CNN structure embedded with CBAM to effectively extract spatial location information but also information of each channel to improving the accuracy of anomaly detection.

$\bullet$ CAINNFlow is a straight-in structure that acts as a distributed converter and can be attached to any feature extractor, which can be any computer vision model using neural networks, such as ResNet, ViT, and ConvNeXt.

$\bullet$ The experimental results using the MVTec \cite{bergmann2019mvtec} AD anomaly detection data set show that the CAINNFlow method we proposed,which indexes with fewer parameters, achieves the advanced level of image-level AUROC(Area Under the Receiver Operating Characteristic) and pixel-level AUROC.

% The Elsevier cas-dc class is based on the
% standard article class and supports almost all of the functionality of
% that class. In addition, it features commands and options to format the
% \begin{itemize} \item document style \item baselineskip \item front
% matter \item keywords and MSC codes \item theorems, definitions and
% proofs \item lables of enumerations \item citation style and labeling.
% \end{itemize}

% This class depends on the following packages
% for its proper functioning:

% \begin{enumerate}
% \itemsep=0pt
% \item {natbib.sty} for citation processing;
% \item {geometry.sty} for margin settings;
% \item {fleqn.clo} for left aligned equations;
% \item {graphicx.sty} for graphics inclusion;
% \item {hyperref.sty} optional packages if hyperlinking is
%   required in the document;
% \end{enumerate}  

% All the above packages are part of any
% standard \LaTeX{} installation.
% Therefore, the users need not be
% bothered about downloading any extra packages.
% 图片1

\section{Related Work}
\subsection{ Anomaly detection method}

Anomaly detection is the most common method in industrial production. One of the most commonly used methods is unsupervised anomaly detection and location, which can effectively detect and locate anomalies  \cite{roth2021towards}. Because in anomaly detection in real life, it is difficult to get a lot of defective sample sample distribution caused by serious imbalance, and the unpredictability of the kinds of causes training focus to include all types of abnormal, made with limited supervision and the application of anomaly detection, unsupervised anomaly detection and location is particularly important. In order to be close to the actual situation, the anomaly check data set of MVTec \cite{bergmann2019mvtec} was used as the main supporting data set in the experiment in this study. It contains $5354$ high-resolution color images of different object and texture classes. It contains normal (that is, defect-free and unlabeled) images for training and abnormal images for testing. Anomalies are represented by more than 70 different types of defects, such as scratches, dents, contamination and various structural changes. This is the first comprehensive multi-target, multi-defect anomaly detection dataset, which is taken from real industrial scenarios and provides pixel-level anomaly annotation.

In addition, during 2017, Y. Feng et al used PCANet to extract both appearance and motion features from 3D gradients on two publicly available datasets, the UCSD Ped1 Dataset and the Avenue Dataset, and video events were automatically represented and modeled in an unsupervised manner. A deep Gaussian Mixture Model (GMM) is constructed using the observed normal events. It is proved that the depth model is effective in detecting abnormal events in video surveillance \cite{feng2017learning}. In 2018, S. Wang et al. modeled the features of ordinary video events on UCSD PED1, PED2, and UMN datasets, and introduced an extreme learning machine (OC-ELM) as the data description algorithm. A new method for automatic detection and location of video anomalies is proposed, which can achieve the most advanced results in both video anomaly detection and location tasks \cite{wang2018video}. In 2019, Nanjun Li et al conducted experiments on three common data sets of UCSD dataset, UMN dataset, and Avenue dataset through Multi-variable Gaussian Full Convolution Adversarial Autoencoder (MGFC-AAE). A video anomaly detection and location method based on deep learning is proposed. \cite{li2019video}. M. Canizo et al. proposed an architecture of supervised multi-time series anomaly detection method based on deep learning, multi-head CNN-RNN(Regressive Neural Network), which combined CNN and RNN in different ways. Different from other methods, we use independent CNN, called convolution head, to handle anomaly detection in multi-sensor systems. This architecture is suitable for multi-time series anomaly detection and has achieved good results in actual industrial scenes \cite{canizo2019multi}. In 2020, X. Zhang et al. proposed a method based on scene perception combining fluid force expression and psychological theory, introduced the line integral convolution flow field visualization technology to segment the moving pedestrians in the scene, and proposed a clustering strategy guided by scene perception to cluster the consistent crowd. Experimental results show that the proposed method achieves higher accuracy in both frames and pixel-level measurement than the existing methods. \cite{zhang2020scene}.
Labiba Gillani Fahad et al. executed a comprehensive activity level assessment of the proposed method on two smart home datasets and used probabilistic neural networks for the classification of presegmented activity instances. It is used to recognize abnormal phenomena in multi-resident activities \cite{fahad2021activity}.
% G. Kim et al. conducted experiments by using NSL-KDD data sets. Firstly, a misuse detection model was established based on the C4.5 decision tree algorithm, and then the model was used to decompose normal training data into smaller subsets. Next, several single-class SVM models are established for the decomposed subset. Experimental results show that the detection rate of objects by this method is superior to that of traditional methods while maintaining a low false-positive rate \cite{kim2014novel}. 
In this study, pixel-accurate Ground Truth(GT) was provided for all anomalies, and the current advanced unsupervised anomaly detection methods were evaluated \cite{yu2021fastflow}, and compared with the CAINNFlow method adopted by us.

\subsection{Feature Extraction}

With the gradual integration of artificial intelligence into people's production and life, deep learning has been widely used in computer vision. Currently, the commonly used methods of feature extraction are respectively based on convolutional Neural Network (CNN) and Transformer. Some studies use ResNet to extract object features. For example, the residual network (ResNet) adopted by Microsoft \cite{li2016demystifying} won the first place in ILSVRC-2015 with an astonishing error rate of $3.6\%$. The 152-layer network they use has lower complexity than the VGG \cite{simonyan2014very} network and solves the degradation problem by introducing residual learning of deep connections. While Transformer has been a big success in Natural Language Processing(NLP), Vision Transformer extends the Transformer model architecture into the realm of computer Vision. Transformer is a good replacement for convolution operations and can still achieve good results in computer vision tasks without relying on convolution. The convolution operation can only consider local feature information, while the attention mechanism in Transformer can comprehensively consider global feature information. In order to be as compatible as possible with Transformer related structures in the NLP domain, Vision Transformer directly migrates Transformer from the NLP domain to the computer Vision domain without changing the Encoder architecture in Transformer as much as possible. Recently, DeiT \cite{touvron2021training}, which is realized by Transformer and distillation technology without any convolution, can get good results only by training on ImageNet with relatively less computation. Later, based on DeiT's work, CaiT \cite{touvron2021going} made Deep Vision Transformer easy to converge and improve accuracy through LayerScale, and CaiT adopted class-attention layers. Make the model more efficient for class token processing. Recently, however, a ConvNeXt \cite{liu2022convnet} based on the CNN model design framework emerged, which surpassed the performance of such models as ViT based on Transformer in terms of overall performance. To sum up, both CNN-based and Transformer Based computer vision models have their own advantages and disadvantages, so the feature extractor modules selected in this study include Transformer Based and CNN Based, both of which are combined with CAINNFlow for experiments.

\subsection{Normalizing Flow}
In recent years, researchers have developed many methods to learn probability distributions of data sets, including Generative Adversarial Networks (gans), Variational Self-encoders (VAE), and Normalizing Flow. However, both GAN and VAE lack accurate evaluation and inference of probability distribution, which leads to poor quality of fuzzy results in VAE, and GAN training also faces problems such as pattern collapse. To overcome the deficiency of GAN and VAE, relevant researchers proposed Normalizing Flow. NF based on Flow can transform random probability distribution into another random probability distribution, while GAN can only transform a random vector into an image, for example, normal distribution into another random distribution including the normal distribution itself. Compared with GAN and VAE, NF can transform the probability distribution corresponding to complex data into a simple probability distribution, for example, the probability distribution corresponding to images in MNIST data into a simple normal distribution, and vice versa, the simple normal distribution can be transformed into complex probability distribution corresponding to images in MNIST.

Therefore, based on the advantages of NF above, we propose CAINNFlow based on NF to achieve distributed transformation on the basis of retaining and effectively extracting spatial structure information.

% The package is available at author resources page at Elsevier
% (\url{http://www.elsevier.com/locate/latex}).
% The class may be moved or copied to a place, usually,\linebreak
% \verb+$TEXMF/tex/latex/elsevier/+, %$%%%%%%%%%%%%%%%%%%%%%%%%%%%%
% or a folder which will be read                   
% by \LaTeX{} during document compilation.  The \TeX{} file
% database needs updation after moving/copying class file.  Usually,
% we use commands like \verb+mktexlsr+ or \verb+texhash+ depending
% upon the distribution and operating system.

\section{Methodology}
\subsection{Definition And Methods of NF}
The general representation of Normalized Flow refers to the gradual approximation of complex distribution to simple Gaussian distribution by using multiple nested reversible functions \cite{yu2021fastflow}. The original distribution has several nested reversible functions and then transforms into any other distribution (including the original distribution itself), where the corresponding increase or decrease in the probability area is the product of the Jacobian determinant of all the reversible functions. Normalizing Flow \cite{cunningham2020normalizing} is essentially a series of reversible functions, so Normalizing Flow is reversible. Meanwhile, the probability density distribution of samples can be converted back to its corresponding original distribution through the reverse process of Normalizing Flow \cite{agnelli2010clustering}.

In simple terms, Normalizing Flow is a set of invertible functions, or the analytic inverse of these functions can be computed.For example, $f(x)=x+2$ is a reversible function because every input has and only one unique output, and vice versa, and $f(x)=x^{2}$ is not a reversible function.Such functions are also called bijective functions.

In particular, I'm given an invertible mapping$f:\mathbb{R}^{d} \xrightarrow{} \mathbb{R}^{d}$, and use it to put random variables $z~q(z)$ Transform to a new variable $z^{'} = f(z)$ After, the distribution of the new variable is:$q\left(x^{\prime}\right)=g(z)\left|\operatorname{Let} \frac{\partial f^{-1}}{\partial x^{2}}\right|=q(z)\left|\operatorname{det} \frac{\partial f}{\partial z}\right|^{-1}$ .Then, in order to build a sufficiently complex distribution, we can design the model structure with multiple similar reversible mappings and nested sequences of mappings:
% \begin{figure*}[htbp]
% \centering
% \includegraphics[scale=1]{pic/Bimage.eps}
% \caption{xxxxx}
% \label{fig1}
% \end{figure*}
\begin{figure*}[]
	\centering
	\includegraphics[width=.95\linewidth]{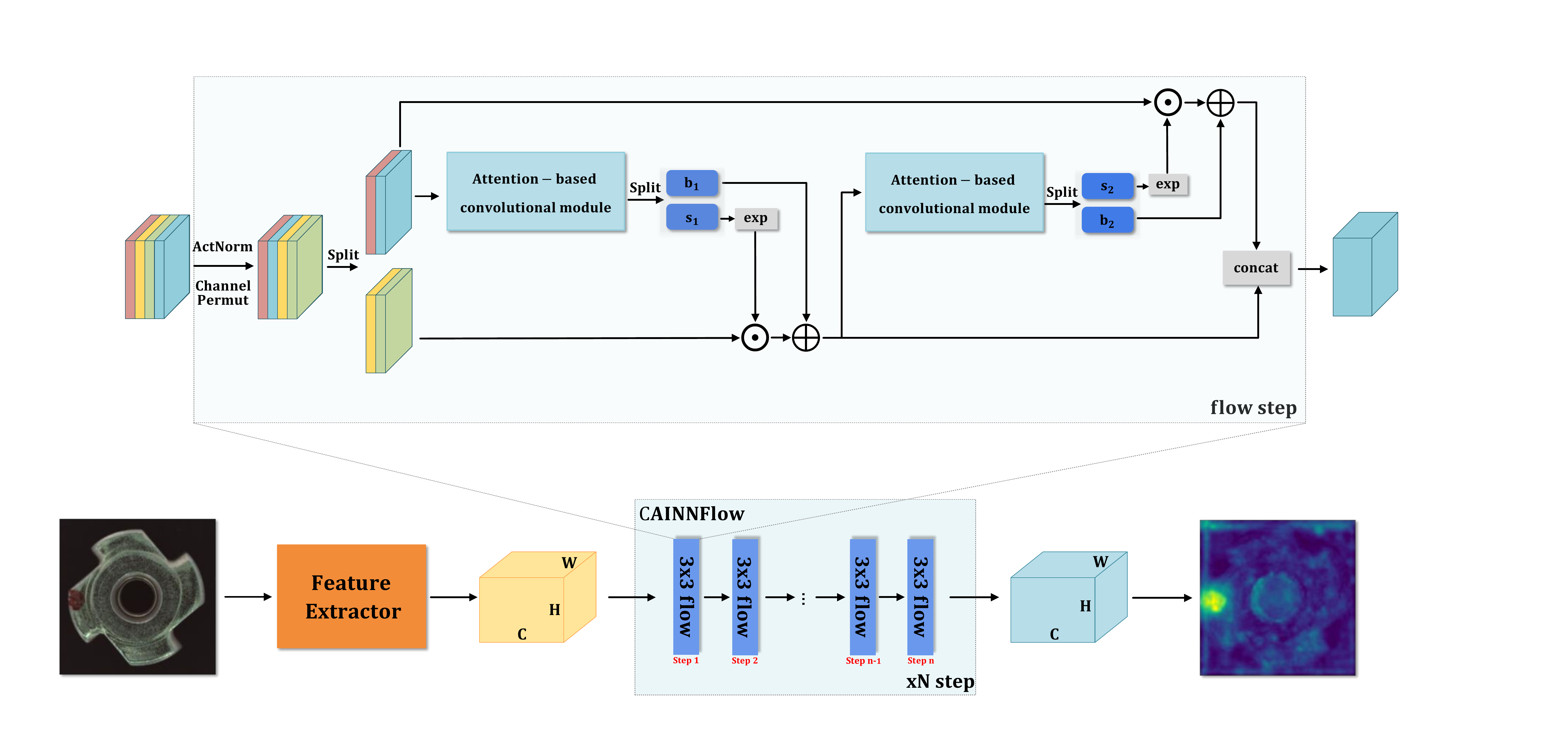}
% 	\includegraphics[width=18cm]{pic/Bimage.eps}
	%\fbox{\rule[-.5cm]{4cm}{4cm} \rule[-.5cm]{4cm}{0cm}}
	\caption{invertible process}
	\label{fig:fig2}
\end{figure*}
% \begin{center}
\begin{equation}
\begin{aligned}
&\mathbf{v}_{1}=\mathbf{u}_{1} \odot \exp \left(s_{2}\left(\mathbf{u}_{2}\right)\right)+t_{2}\left(\mathbf{u}_{2}\right) \\
&\mathbf{v}_{2}=\mathbf{u}_{2} \odot \exp \left(s_{1}\left(\mathbf{v}_{1}\right)\right)+t_{1}\left(\mathbf{v}_{1}\right)
\end{aligned}
\end{equation}
% \end{center}
In the figure: \ref{fig:fig2}, the basic building block of a reversible neural network is the affine coupling layer generalized by the Real NVP model. Its working principle is to divide the input data into two parts $u_{1}$ and $u_{2}$, which are converted by learning functions $s_{i}(u_{i})$ and $t_{i}(u_{i})$ and coupled alternately.

The specific operation is as follows: firstly, input data $u_{2}$ is substituted into function $s_{2}$, the result of $s_{2} (u_{2})$ is substituted into base E for exponential operation, and then the result of input data $u_{1}$ and $exp (s_{2} (u_{2}))$ are dot multiplied to get $u_{1} \bigodot$ After $exp (s_{2} (u_{2}))$, add the result with the result of input data $u_{2}$ into the function $T_{2}$, and finally get our output node $v_{1}$. Similarly, we can perform a series of operations to get the output node $v_{2}$.

\begin{figure}[h]
	\centering
	\includegraphics[width=.95\linewidth]{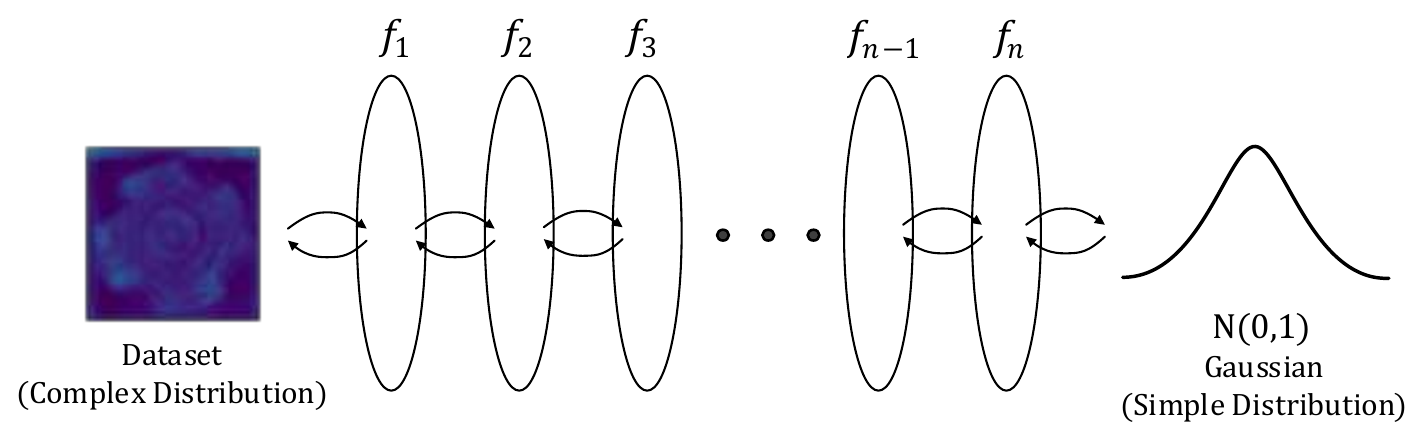}
	%\fbox{\rule[-.5cm]{4cm}{4cm} \rule[-.5cm]{4cm}{0cm}}
	\caption{Probability transformation process}
	\label{fig:fig12}
\end{figure}

% \begin{center}
\begin{equation}
\mathbf{z}_{K}=f_{K} \circ \ldots \circ f_{2} \circ f_{1}\left(\mathbf{z}_{0}\right)
\end{equation}
% \end{center}
In this way,$\ln q_{K}\left(\mathbf{z}_{K}\right)=\ln q_{0}\left(\mathbf{z}_{0}\right)-\sum_{k=1}^{K} \ln \left|\operatorname{det} \frac{\partial f_{k}}{\partial z_{k-1}}\right|$ When you calculate the distribution after the transformation $q_{k}$. Value is not evaluated explicitly $q_{k}$ But by the initial distribution $q_{0}$. Through the above process, NF can transform the original distribution into a new distribution with a series of reversible mappings. By optimizing the variable parameters of the bijective function during training, the bijective function can transform the basic distribution into an arbitrary distribution. Each bijective function can be written as a layer of a network, with the optimizer learning the parameters and finally fitting the real data. By using the maximum likelihood estimation method, the distribution problem of fitting real data is changed into the log-likelihood problem of the probability after fitting transformation, and the log-likelihood problem is also used for the stability of calculation. The traditional Normalized Flow has the following characteristics \cite{kobyzev2020normalizing} :

$\bullet$ A mapping from input to output is bijective, that is, its inverse function exists.
    
$\bullet$ Both forward and reverse mappings are effectively computable.
    
$\bullet$ The map has a tractable Jacobian determinant, so the probabilities can be explicitly converted by variable formulas.
    
The proposal of NF provides an effective idea and method to solve the fitting problem of complex distribution. Reversible functions can map the vector space in the Cartesian coordinate system into different vector Spaces and reversely map back to the original vector space using the inverse operation of reversible functions. It is worth noting that the area of the vector space is changed in the above mapping process by the Jacobian determinant of the reversible function.

Since some information is lost in the forward process, an additional potential output variable $z$ is introduced, which is trained to capture information related to $x$ but not contained in $y$. In addition, the training network needs to adjust $p(z)$ according to the Gaussian distribution. Namely, $p (x | y)$ is adjusted to a certain function $x = g (y, z)$, this function will be known to the distribution of $p (z)$ in the case of meet $y$ to $x$ space.

If $x \in RD$ $y \in RM$, then due to the loss of information in the forward process, the inherent dimension m of $y$ must be less than $D$, even though $M$ may be greater than $D$. We hope that based on the model $q (x | y)$ to predict the $\rho (x | y)$, with the introduction of implicit variable $z$ and $q (x | y)$ can be in $g (y, z;\theta)$ appears as follows:

\begin{equation}
\begin{aligned}
\mathbf{x}=g(\mathbf{y}, \mathbf{z} ; \theta) \quad \text { with } \quad \mathbf{z} \sim p(\mathbf{z})=\mathcal{N}\left(\mathbf{z} ; 0, I_{K}\right)
\end{aligned}    
\end{equation}
Correspondingly, the forward process can also be expressed by $f(x;\Theta)$ said:  
\begin{equation}
\begin{aligned}
\lceil\mathbf{y}, \mathbf{z}]=f(\mathbf{x} ; \theta) &=\left[f_{\mathbf{y}}(\mathbf{x} ; \theta), f_{\mathbf{z}}(\mathbf{x} ; \theta)\right]=g^{-1}(\mathbf{x} ; \theta) \\
f_{\mathbf{y}}(\mathbf{x} ; \theta) & \approx s(\mathbf{x})
\end{aligned}
\end{equation}
Bidirectional training $f$ and $G$ can avoid the problems in GAN and Bayesian neural networks. Since INN requires $f = G-1$, the dimensions of both sides (whether inherent or displayed) should be the same, requiring the dimension of variable $Z$, $K = D -- m$. If it causes $M+K > D$, we need to complement the $x$ vector with the $0$ vector of the $M+ k-d$ dimensions.

Combination of all these definitions, our network will be $q (x | y)$ is expressed as:
% \begin{center}
\begin{equation}
\begin{aligned}
q(\mathbf{x} &=g(\mathbf{y}, \mathbf{z} ; \theta) \mid \mathbf{y}) =p(\mathbf{z})\left|J_{\mathbf{x}}\right|^{-1}\\
% with Jacobian determinant
J_{\mathbf{x}}&=\operatorname{det}\left(\left.\frac{\partial g(\mathbf{y}, \mathbf{z} ; \theta)}{\partial[\mathbf{y}, \mathbf{z}]}\right|_{\mathbf{y}, f_{\mathbf{z}}(\mathbf{x})}\right)
\end{aligned}
\end{equation}
% \end{center}

    By alternating forward and backward iterations and accumulating bidirectional gradients before parameters are updated, the input and output errors are reduced and training is more efficient.

$s_{i}(u_{i})$ and $t_{i}(u_{i})$ subnets (complex functions) in NF are usually multi-layer perceptrons, which need to flatten and extrude the input visual features from $2D$ to $1D$, thus destroying the spatial position relationship in the feature graph and losing the spatial structure information. To retain and effectively extract spatial structure information, we designed a complex function model with alternating CBAM embedded in the stacked $3×3$ full convolution in this study, which can retain and effectively extract spatial structure information in the Normalized Flow model. Its model structure is the combination of two-dimensional convolution and CBAM structure. CBAM is a convolution attention module that integrates channel attention and spatial attention mechanism, and the tensor shape of input and output of CBAM is consistent so that it can be seamlessly integrated into the structure constituted by CNN. Therefore, we propose a structure named CAINNFlow that can retain spatial information and extract spatial information effectively.

As the characteristic distribution of abnormal samples is different from that of normal samples, the likelihood value of abnormal samples obtained by CAINNFlow should be lower than that of normal samples. Therefore, the likelihood value of test samples can be used as the anomaly score to detect and locate anomalies.

\subsection{CBAM}
% Typically, the forward process from parameters to measurement space is a well-defined function, while the inverse problem is ambiguous because one measurement can map to multiple different sets of parameters. There is one type of neural network suitable for solving this kind of problem - the inverse neural network INN. Whereas traditional neural networks try to solve this type of problem directly, INNs can learn it along with a well-defined forward process, using additional implicit output variables to capture information that might be lost in the forward process.
% \subsubsection{CBAM}
\begin{figure}[h]
	\centering
	\includegraphics[width=.95\linewidth]{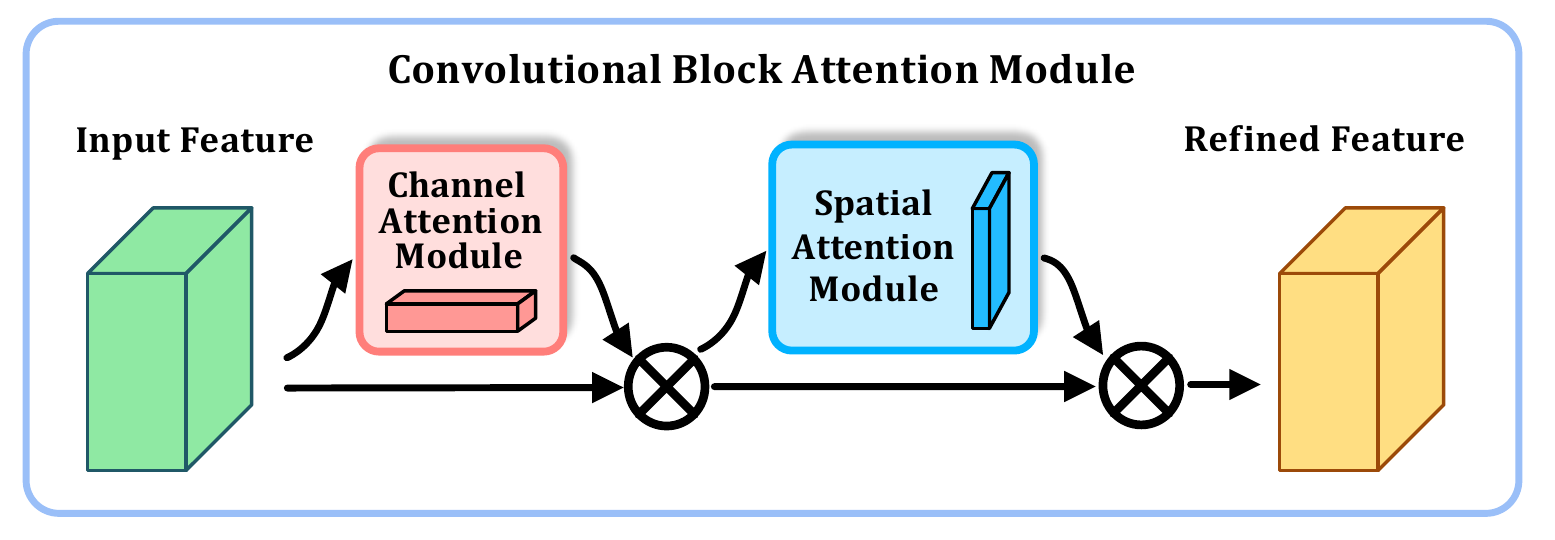}
	%\fbox{\rule[-.5cm]{4cm}{4cm} \rule[-.5cm]{4cm}{0cm}}
	\caption{The overview with CBAM}
	\label{fig:fig3}
\end{figure}
CBAM is a convolution attention module proposed by Sanghyun et al., which integrates channel attention and spatial attention mechanism \cite{woo2018cbam}. CBAM can be seamlessly integrated into CNNs and can conduct end-to-end training with CNNs.

The structure of CBAM is shown in the figure: \ref{fig:fig3} below, which has two sub-modules: channel attention and spatial attention. Place the two modules sequentially in CBAM. Given a feature graph $F \in R^{C×H×W}$, input it into the channel attention module will generate channel attention diagram $Mc \in R^{C×1×1}$, which can be obtained by multiplying with $F$ element $F^{'} \in R^{C * H * W}$. Then as the input, the two-dimensional spatial attention graph $Ms \in R^{1×H×W}$ is generated through the spatial attention module, and then the sum is made the elements are multiplied to get the final output.It is worth noting that we use a cascade structure of "channel attention first, space attention second", and this stacked topology performs better in other experiments. The calculation process of CBAM is as follows:

\begin{equation}
\begin{aligned}
\mathrm{F}^{\prime} &=\mathrm{M}_{\mathrm{c}}(\mathrm{F}) \otimes \mathrm{F},\\
\mathrm{F}^{\prime \prime} &=\mathrm{M}_{\mathrm{s}}(\mathrm{F}^{\prime}) \otimes \mathrm{F}^{\prime}
\end{aligned}
\end{equation}

where $\otimes$denotes element-wise multiplication. During multiplication, the attention values are broadcasted (copied) accordingly: channel attention values are broadcasted along the spatial dimension, and vice versa. $F^{''}$ is the final refined output.
\begin{figure}[h]
	\centering
	\includegraphics[width=.95\linewidth]{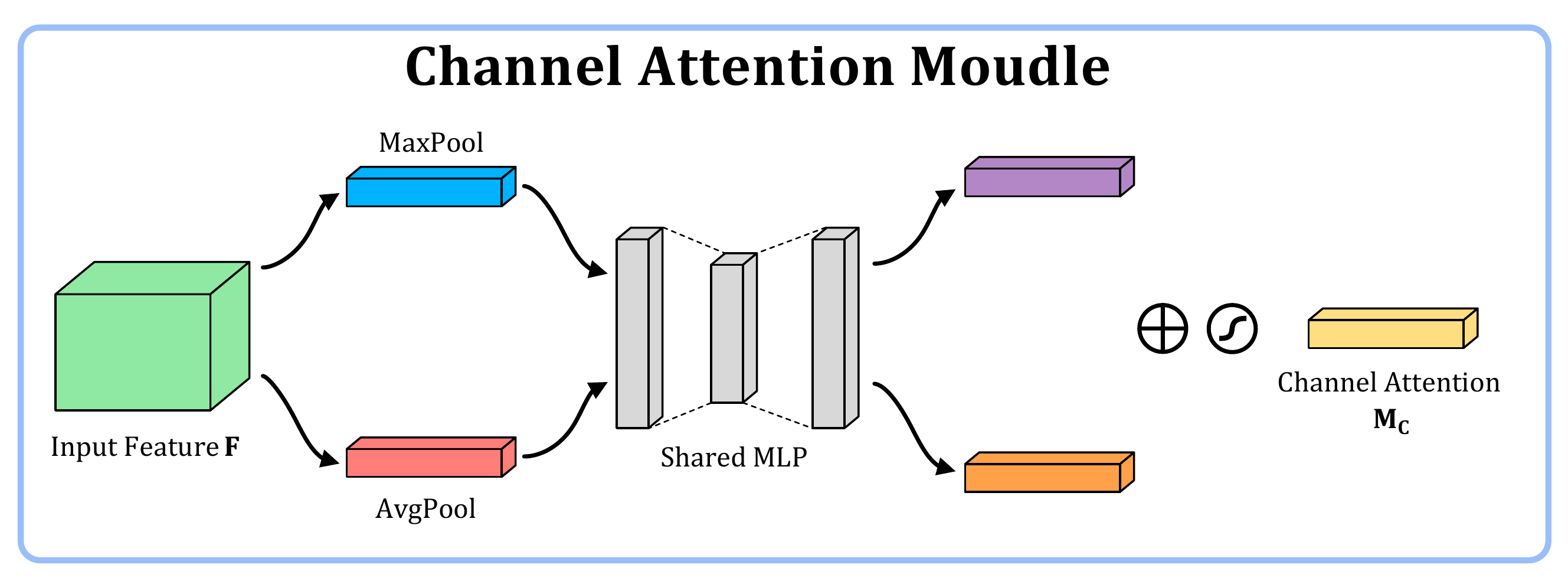}
	%\fbox{\rule[-.5cm]{4cm}{4cm} \rule[-.5cm]{4cm}{0cm}}
	\caption{Channel Attention Module}
	\label{fig:fig4}
\end{figure}
The structure of the channel attention module is shown in the figure: \ref{fig:fig4}.In the channel attention module, the maximum pooling and average pooling operations are used to compress each channel, so that the information on the input feature graph $F$ is aggregated into the feature graph $F^{c}_{avg}\in R^{C×1×1}$ and $F^{c}_{max}\in R^{C×1×1}$. $F^{c}_{avg}$ and $F^{c}_{max}$ represent Average-pooled features and max-pooled Features, respectively. The two feature maps are then input into a shared network of a multi-layer perceptron. To reduce parameter overhead, the hidden activation size is set to $R^{C/r×1×1}$, where R is the reduction ratio. Add the elements of the two outputs of the shared network and use the Sigmoid activation function to generate our channel. Note the graph $Mc\in R^{C×1×1}$. The calculation process of channel attention is as follows:
\begin{equation}
\begin{aligned}
\mathbf{M}_{c}(\mathbf{F}) &=\sigma(\operatorname{MLP}(\operatorname{Avg} \text { Pool }(\mathbf{F}))+\operatorname{MLP}(\operatorname{Max} \text { Pool }(\mathbf{F}))) \\
&=\sigma\left(\mathbf{W}_{\mathbf{1}}\left(\mathbf{W}_{0}\left(\mathbf{F}_{\text {avg }}^{\mathrm{c}}\right)\right)+\mathbf{W}_{\mathbf{1}}\left(\mathbf{W}_{0}\left(\mathbf{F}_{\max }^{\mathrm{c}}\right)\right)\right)
\end{aligned}
\end{equation}

where $\sigma$ denotes the sigmoid function, $W_0\in R^{C/r×C}$, and $W_1\in R^{C×C/r}$. Note that the MLP weights, $W_0$and $W_1$, are shared for both inputs and the ReLU activation function is followed by $W_0$.
\begin{figure}[h]
	\centering
	\includegraphics[width=.95\linewidth]{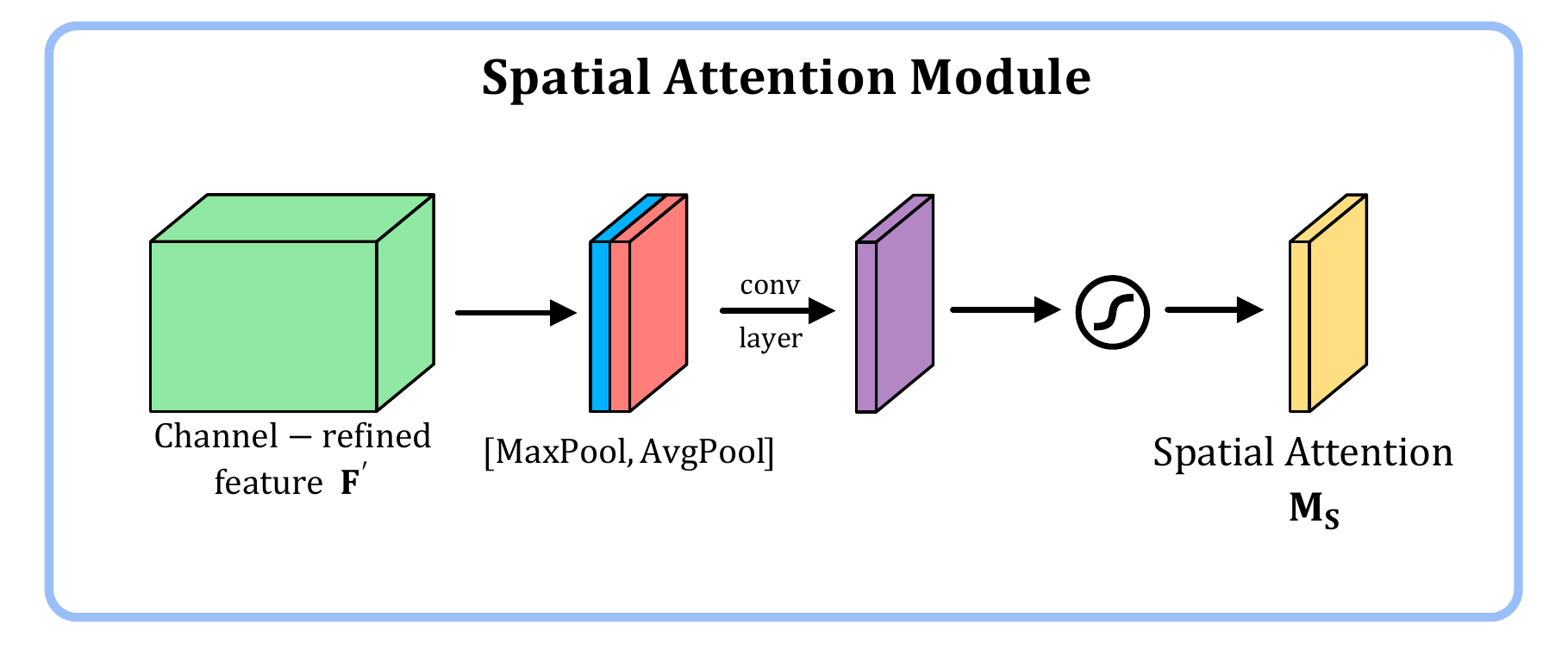}
	%\fbox{\rule[-.5cm]{4cm}{4cm} \rule[-.5cm]{4cm}{0cm}}
	\caption{Spatial Attention Module}
	\label{fig:fig5}
\end{figure}
The structure of the spatial attention module is shown in the figure: \ref{fig:fig5}. In the spatial attention module, we first perform mean pooling and max pooling operations on the input feature map F along the channel axis to aggregate the information on the feature map $F$ into the feature maps $F^{s}_{avg
}\in R^{1×H×W}$and $F^{s}_{max} \in R^{1×H×W}$, $F^{s}_{avg}$ and $F^{s}_{max}$ represent average-pooled features and max-pooled features of all channels, respectively. The two feature maps are then concatenated and fed into a standard convolutional layer for convolution. Finally, the sigmoid operation is performed on the output of the convolutional layer to generate a two-dimensional spatial attention map $Ms(F) \in R^{H×W}$. The calculation process of spatial attention is:
\begin{center}
\begin{equation}
\begin{aligned}
\mathbf{M}_{s}(\mathbf{F}) &=\sigma\left(f^{7 \times 7}([\operatorname{AvgPool}(\mathbf{F}) ; \text { MaxPool }(\mathbf{F})])\right) \\
&=\sigma\left(f^{7 \times 7}\left(\left[\mathbf{F}_{\text {avg }}^{s} ; \mathbf{F}_{\text {max }}^{s}\right]\right)\right)
\end{aligned}
\end{equation}
\end{center}

where $\sigma$ denotes the sigmoid function and $f^{7×7}$represents a convolution operation with the filter size of $7×7$.

% Figure 1: The overview with CBAM

% The author names and affiliations could be formatted in two ways:
% \begin{enumerate}[(1)]
% \item Group the authors per affiliation.
% \item Use footnotes to indicate the affiliations.
% \end{enumerate}
% See the front matter of this document for examples. 
% You are recommended to conform your choice to the journal you 
% are submitting to.
\subsection{Comparison of different spatial information extraction structures of CAINNFlow}
\begin{figure}[h]
	\centering
	\includegraphics[width=.95\linewidth]{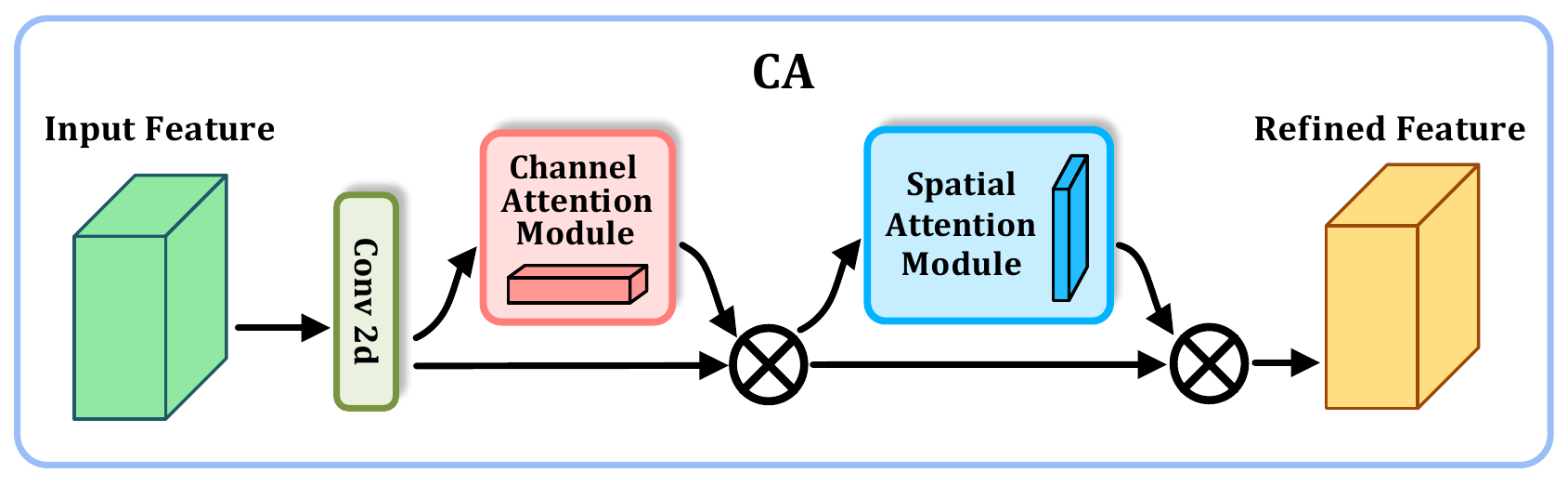}
	%\fbox{\rule[-.5cm]{4cm}{4cm} \rule[-.5cm]{4cm}{0cm}}
	\caption{CA Module}
	\label{fig:fig6}
\end{figure}

CAINNFlow has three modules to extract spatial information: CA, AC, and CAC, as well as a CC module for ablation experiments. The structure of the CA module is connected to a CBAM after a $3×3$ convolution layer, as shown in the figure: \ref{fig:fig6}. The final output of the input feature graph $F$ is obtained through CBAM after convolution. The calculation process of the CA module is as follows:
\begin{equation}
\begin{aligned}
F^{\prime}&=\operatorname{Con}(F) \\
F^{\prime \prime}&=M c\left(F^{\prime}\right) \otimes F^{\prime} \downarrow \\
F^{\prime \prime \prime}&=M s\left(F^{\prime \prime}\right) \otimes F^{\prime \prime}
\end{aligned}
\end{equation}

\begin{figure}[h]
	\centering
	\includegraphics[width=.95\linewidth]{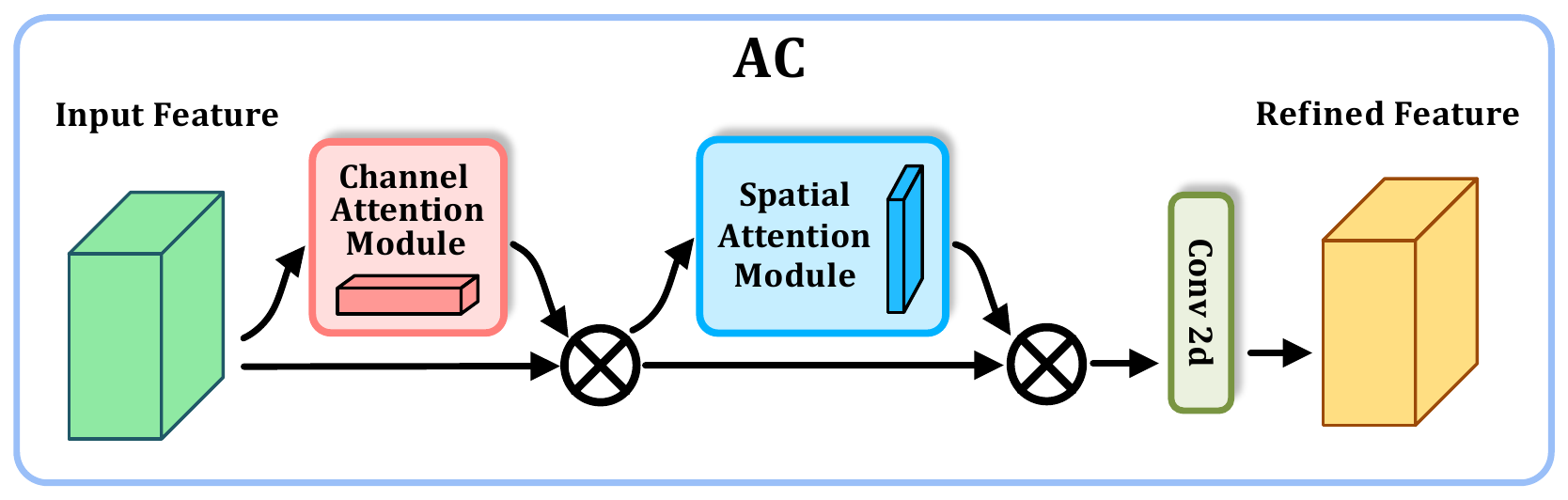}
	%\fbox{\rule[-.5cm]{4cm}{4cm} \rule[-.5cm]{4cm}{0cm}}
	\caption{AC Module}
	\label{fig:fig7}
\end{figure}

Among them, $\otimes$ For element-level multiplication, Con for convolution operation, $F^{'}$ Represents the output of the convolution. $F^{''}$ Note the output of the graph elements multiplied by each other,$F^{'''}$ for $F^{''}$, And the final output of the multiplication of the force diagram elements in two-dimensional space.
The structure of the AC module is connected with a $3×3$ convolution layer after a CBAM, and its structure diagram is shown in the figure: \ref{fig:fig7}.The final output of input feature graph $F$ is obtained after CBAM and convolution.The AC module is calculated as follows:
\begin{equation}
\begin{aligned}
F^{\prime}&=\mathrm{M}_{c}(F) \otimes F^{\prime} \\
F^{\prime \prime}&=M_{s}\left(F^{\prime}\right) \otimes F^{\prime} \downarrow \\
F^{\prime \prime \prime}&=\operatorname{Con}\left(F^{\prime \prime}\right)
\end{aligned}
\end{equation}

\begin{figure}[h]
	\centering
	\includegraphics[width=.95\linewidth]{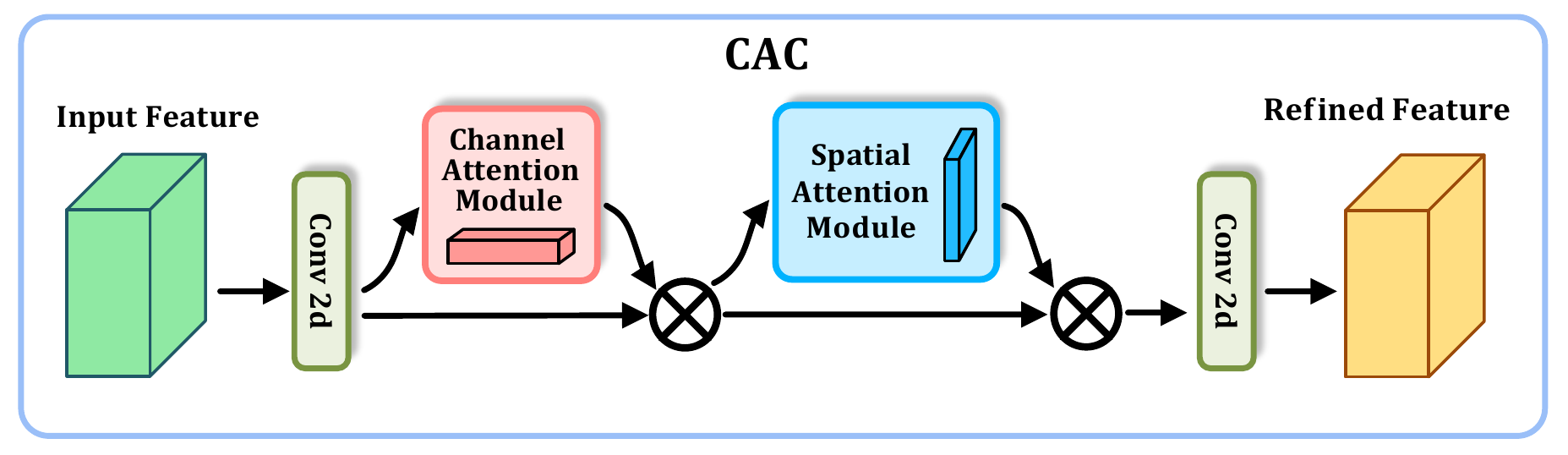}
	%\fbox{\rule[-.5cm]{4cm}{4cm} \rule[-.5cm]{4cm}{0cm}}
	\caption{CAC Module}
	\label{fig:fig8}
\end{figure}

Among them $\otimes$ For element-level multiplication, and Con for convolution operation. $F^{'}$ for $F$ Notice the output when you multiply the graph elements, $F^{''}$ For $F^{'}$ the  And the output of the multiplication of the force diagram elements in two-dimensional space, $F^{'''}$ for $F^{''}$ The final output after convolution.
The structure of the CAC module is that a $3×3$ convolution layer is followed by a CBAM, and then another $3×3$ convolution layer, as shown in the figure: \ref{fig:fig8}. Input feature graph $F$ is first convolved and then input CBAM, and then the output of CBAM is convolved to obtain the final output. The calculation process of the CAC module is as follows:
\begin{equation}
\begin{aligned}
F^{\prime}&=\operatorname{Con}(F)_{\downarrow} \\
F^{\prime \prime}&=M c\left(F^{\prime}\right) \otimes F_{\downarrow}^{\prime} \\
F^{\prime \prime \prime}&=M s\left(F^{\prime \prime}\right) \otimes F_{\downarrow}^{\prime} \\
F^{(4)}&=\operatorname{Con}\left(F^{\prime \prime \prime}\right)
\end{aligned}
\end{equation}

Among them $\otimes$ For element-level multiplication, Conv stands for convolution operation.$F^{'}$ Is the output after the first convolution, $F^{''}$ for $F^{'}$ Notice the output when you multiply the graph elements, $F^{'''}$ for $F^{''}$ And the output of the multiplication of the force diagram elements in two-dimensional space, $F^{(4)}$ for $F^{'''}$ The final output after the second convolution.

\begin{figure}[h]
	\centering
	\includegraphics[width=.95\linewidth]{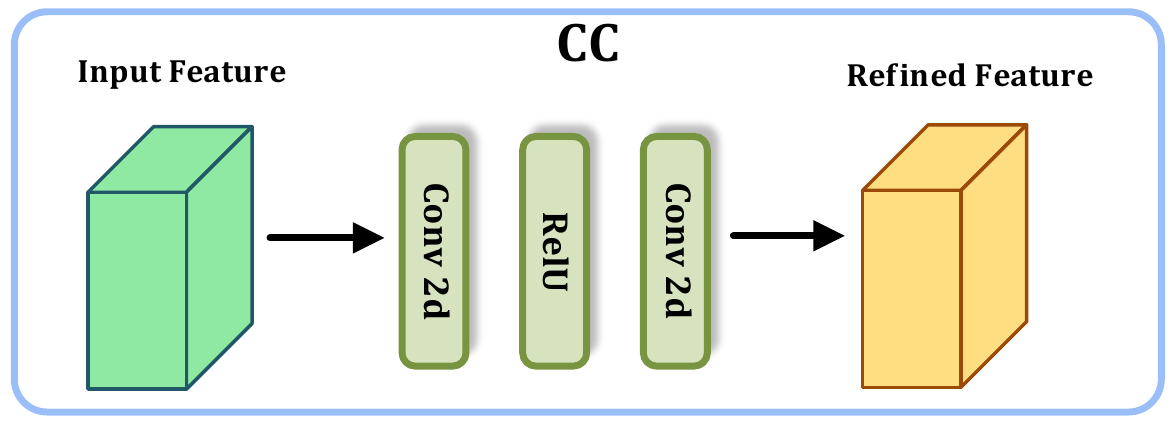}
	%\fbox{\rule[-.5cm]{4cm}{4cm} \rule[-.5cm]{4cm}{0cm}}
	\caption{CC Module}
	\label{fig:fig9}
\end{figure}

The structure of the CC module is two $3×3$ convolution layers, and its structure diagram is shown in the figure: \ref{fig:fig9}. The final output of the input feature graph F is obtained after two levels of convolution. The calculation process of the CC module is as follows:
\begin{equation}
\begin{aligned}
F^{\prime}&=\operatorname{Con}(F) \\
F^{\prime \prime}&=\operatorname{Con}(F)^{\prime} \\
\end{aligned}
\end{equation}

Conv stands for convolution operation.$F^{'}$ Is the output after the first convolution and the $F^{''}$ output after the second convolution.
% \subsection{Feature extraction}
% In terms of feature extraction research for anomaly detected objects, many current studies try to preserve the spatial structure of images, and some progress has been made. However, the feature extraction ability is insufficient when dealing with large axis tensors due to its simple structure. So throughout the research process, we first extract representative features from the input image through a visual transformer to distinguish these anomalous regions from other local regions. For ResNet11, ResNet solves the degradation problem of deep networks through residual learning, allowing us to train deeper networks, we directly use the features of the last layer in the first three blocks and put these features into the three corresponding Flow models.

% \subsection{summary}
% We achieve this by designing a two-dimensional flow model Flow with a lightweight structure and a standard normal distribution for projecting the feature distribution of normal images into training and being able to use probabilities as anomaly scores in testing. For unsupervised anomaly detection and localization, it outperforms state-of-the-art methods in terms of accuracy and detection efficiency.
\section{Experiments}
\subsection{The data set}
  
The detection of abnormal structures in natural image data is very important for many types of research in computer vision. The data set used in this study is the MVTec Anomaly Detection dataset, which contains $5354$ high-resolution color images of different object and texture categories, as well as normal images for training without defects and defective anomaly images for testing. The abnormal images show more than 70 different types of defects, such as scratches, dents, contamination, and various structural changes.

This dataset is a new unsupervised anomaly detection dataset that simulates real industrial detection scenarios and offers the possibility of evaluating unsupervised anomaly detection methods for various textures and object classes with different types of exceptions. Since this dataset provides pixel-level accurate GT annotation for anomaly areas in the image, we can evaluate anomaly detection methods for image-level classification and pixel-level segmentation. At the same time, the proposed data set promotes the development of new unsupervised anomaly detection methods. The development of unsupervised anomaly detection methods requires data to train and evaluate new methods and ideas, so in an unsupervised setting, we train our model with normal images for each category and evaluate it in test images that contain both normal and abnormal images.

\begin{table*}[]
\resizebox{\textwidth}{!}{
\begin{tabular}{|
>{\columncolor[HTML]{DAE8FC}}c |c|c|c|l|c|c|c|c|c|}
\hline
Method     & \cellcolor[HTML]{DAE8FC}PatchSVDD & \cellcolor[HTML]{DAE8FC}SPADE & \cellcolor[HTML]{DAE8FC}DifferNet & \multicolumn{1}{c|}{\cellcolor[HTML]{DAE8FC}PaDiM} & \cellcolor[HTML]{DAE8FC}Cut Paste & \cellcolor[HTML]{DAE8FC}PatchCore & \cellcolor[HTML]{DAE8FC}CFlow & \cellcolor[HTML]{DAE8FC}CC(3×3) & \cellcolor[HTML]{DAE8FC}CAINNFlow(3×3) \\ \hline
carpet     & (92.64,   92.41)                  & (98.62,   97.53)              & (83.96,   -)                      & (-   98.92)                                        & (\textbf{100.00},   98.31)                  & (98.34,   98.62)                  & (\textbf{100.00},   99.18)             & (\textbf{100.00},   99.28)               & (\textbf{100.00},   \textbf{99.40})                     \\ \hline
grid       & (94.63,   96.17)                  & (98.89,   93.57)              & (97.06,   -)                      & (-   97.26)                                        & (96.21,   97.52)                  & (98.00,   98.67)                  & (97.36,   98.86)              & (99.90,   98.58)                & (\textbf{100.00}, \textbf{98.88})                       \\ \hline
leather    & (90.92,   97.41)                  & (99.16,   97.37)              & (99.42,   -)                      & (-   99.22)                                        & (95.05,   99.34)                  & (\textbf{100.00},   99.29)                 & (97.44,   99.60)              & (\textbf{100.00},   99.58)               & (\textbf{100.00},   \textbf{99.68})                     \\ \hline
tile       & (97.79,   91.44)                  & (89.75,   87.43)              & (92.73,   -)                      & (-   94.07)                                        & (\textbf{100.00},   90.53)                 & (98.71,   95.62)                  & (96.74,   97.53)              & (\textbf{100.00},   97.16)               & (\textbf{100.00}, \textbf{97.54})                       \\ \hline
wood       & (96.76, 91.07)                    & (95.84, 88.47)                & (99.83, -)                        & (- 94.87)                                          & (98.75, 95.40)                    & (99.22, 95.04)                    & (99.52, \textbf{96.66})                & (\textbf{99.88}, 95.90)                  & (99.78, 95.51)                        \\ \hline
bottle     & (98.56,   98.05)                  & (98.03,   98.11)              & (99.14,   -)                      & (-   98.28)                                        & (\textbf{100.00},   97.73)                 & (99.92,   98.13)                  & (99.56,   98.26)              & (\textbf{100.00},   98.08)               & (\textbf{100.00},   \textbf{98.49})                     \\ \hline
cable      & (90.32,   96.76)                  & (93.21,   97.22)              & (86.94,   -)                      & (-   96.73)                                        & (\textbf{100.00},   90.24)                 & (99.12,   98.05)                  & (\textbf{100.00},   97.61)             & (\textbf{100.00},   98.06)                & (\textbf{100.00},   \textbf{98.71})                     \\ \hline
capsule    & (76.37, 95.58)                    & (98.63, \textbf{99.02})                & (88.58, -)                        & (- 98.49)                                          & (98.87, 97.61)                    & (98.13, 98.82)                    & (99.30, 98.97)                & (\textbf{100.00}, 99.00)                 & (99.96, 98.94)                        \\ \hline
hazelnut   & (92.03,   97.52)                  & (98.67,   99.02)              & (99.14,   -)                      & (-   98.09)                                        & (93.25,   97.28)                  & (99.67,   98.48)                  & (96.81,   98.94)              & (\textbf{100.00},   99.18)                & (\textbf{100.00},   \textbf{99.29})                     \\ \hline
meta nut   & (93.81,   97.92)                  & (96.92,   98.13)              & (95.19,   -)                      & (-   97.21)                                        & (86.49,   93.32)                  & (\textbf{100.00},   98.27)                 & (91.62,   98.56)              & (99.94,   98.35)                & (\textbf{100.00},   \textbf{99.11})                     \\ \hline
pill       & (86.11,   95.07)                  & (96.77,   96.21)              & (95.94,   -)                      & (-   95.86)                                        & (\textbf{99.76},   95.66)                  & (96.62,   97.13)                  & (98.82,   97.92)              & (99.18,   \textbf{98.97})                & (98.81, 98.51)                        \\ \hline
screw      & (81.23,   95.62)                  & (99.54,   98.93)              & (99.12,   -)                      & (-   98.46)                                        & (90.70,   96.66)                  & (98.06,   99.29)                  & (\textbf{99.72},   98.92)              & (97.74,   99.30)                & (98.72, \textbf{99.68})                        \\ \hline
toothbrush & (\textbf{100.00},   98.10)                 & (98.63,   97.80)              & (96.13,   -)                      & (-   98.64)                                        & (97.53,   98.11)                  & (\textbf{100.00},   98.66)                 & (95.47,   99.18)              & (96.44,   99.16)                & (97.23,   \textbf{99.55})                      \\ \hline
transistor & (91.64,   97.12)                  & (80.98,   94.11)              & (96.07,   -)                      & (-   97.40)                                        & (\textbf{100.00},   93.18)                 & (\textbf{100.00},   96.26)                 & (98.61,   97.51)              & (99.76,   97.28)                & (\textbf{100.00},   \textbf{97.58})                     \\ \hline
zipper     & (97.89,   95.14)                  & (98.78,   96.53)              & (98.56,   -)                      & (-   98.44)                                        & (99.57,   98.67)                  & (98.67,   98.60)                  & (98.13,   98.63)              & (99.42,   98.58)                & (\textbf{99.60}, \textbf{98.71})                        \\ \hline
AUCROC     & (92.05,   95.69)                  & (96.16,   95.96)              & (95.19,   -)                      & (-,97.46)                                          & (97.08,   95.97)                  & (98.96,   97.93)                  & (97.94,   98.42)              & (99.48,   98.43)                & (\textbf{99.61}, \textbf{98.64})                        \\ \hline
\end{tabular}}
\caption{}
\label{tab1}
\end{table*}

\begin{figure*}[h]
	\centering
	\includegraphics[width=.95\linewidth]{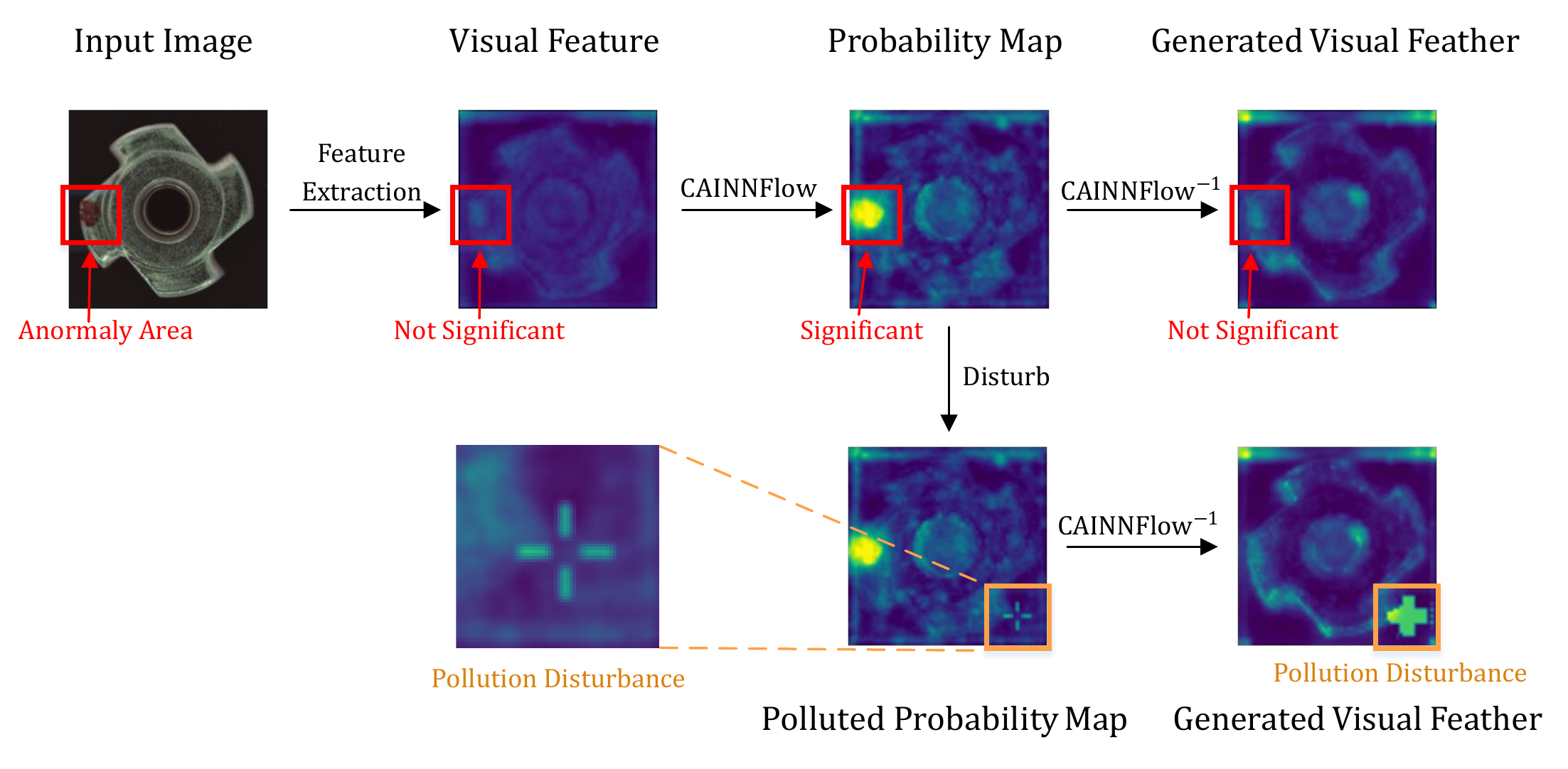}
	%\fbox{\rule[-.5cm]{4cm}{4cm} \rule[-.5cm]{4cm}{0cm}}
	\caption{Feature Visualization and Generation}
	\label{fig:fig10}
\end{figure*}
\subsection{The evaluation}
  
In this study, the method we used and other methods compared with it need to be measured by the image or pixel-level region under the receiver Operation Characteristic Curve (AUROC) during the comparison of their respective performance. AUROC is an indicator used to measure the performance of a classifier. It reflects the performance of the classifier through the area between the receiver operating characteristic curve and the coordinate axis. Its significance can be understood as the expectation that uniformly selected random positive samples (positive samples) rank ahead of uniformly selected random negative samples (negative samples). AUROC is a value between $0$ and $1$. When the AUROC value is close to $1$, it indicates that the classifier has a better classification of positive and negative samples. In the locating task, we need to output an exception score for each pixel for comparison. Meanwhile, AUROC is also a dichotomous evaluation method commonly used in machine learning, which directly means the area under the ROC curve. A pair of samples (one positive sample and one negative sample) are randomly selected, and then the trained classifier is used to predict the two samples. The probability of a positive sample is predicted to be better than the probability of a negative sample.

There is how to calculate AUROC:
    
        In a data set with $M$ positive samples and $N$ negative samples. There are $M$ times $N$ pairs of samples. So let's figure out how many of these $M$ times $N$ samples, the positive sample is more likely to predict than the negative sample.
\begin{equation}
A U C=\frac{\sum I\left(P_{\text {Positive samples }}, P_{\text {Negative samples }}\right)}{M * N}
\end{equation}

%     Method 2:
    
%         Using the formula:
    
%         1. Rank the predicted probabilities from highest to lowest.
    
%         2. Assign a rank to each probability (rank n for highest probability, n-1 for second-highest probability).
    
%         3. The rank represents the number of samples that the score (probability of prediction) exceeds.  
    
% To find that the score value of the positive sample in the combination is greater than that of the negative sample if the score value of all the positive samples is greater than that of the negative sample, then the score value of the first digit will be larger when it is combined with anyone. We take its rank value as N. However, m-1 in n-1 is the combination of positive samples and positive samples, which is not within the statistical scope (we take N groups for convenient calculation, and there are M corresponding inconsistencies), so we need to subtract.

% Similarly, m-1 of n-1 in the second place is not satisfied, and so on. Therefore, the following formula M*(M+1)/2 is obtained. We can verify that under the assumption that the score of positive samples is greater than that of negative samples, the value of AUC is 1.

%         4. Divided by M * N
% \begin{equation}
% A U C=\frac{\sum_{\text {i$\in$ positiveClass }} rank_{i}-\frac{M(1+M)}{2}}{M \times N}
% \end{equation}
In addition, the advantages of using AUROC to measure the performance of the classifier are as follows:

    $\bullet$ It is not affected by category imbalance, and different sample proportions will not affect the evaluation results of AUC.
    
    $\bullet$ AUC can be directly used as a loss function during training.

Therefore, two evaluation indexes, image-level AUROC and pixel-level AUROC are adopted in anomaly detection and location tasks. Image-level AUROC is an evaluation indicator used to classify whether the whole image is abnormal, and pixel-level AUROC is an evaluation indicator used to judge whether the model accurately classifies whether a single pixel is defective.
 
\begin{table*}[t]
\resizebox{\textwidth}{!}{
\begin{tabular}{|c|c|c|c|c|c|c|c|c|c|c|c|c|c|c|c|}
\hline
\rowcolor[HTML]{DAE8FC} 
\multicolumn{1}{|l|}{\cellcolor[HTML]{DAE8FC}} & bottle & cable & capsule & carpet & hazelnut & leather & metal nut & pill  & tile  & toothbrush & transistor & wood  & grid  & screw & zipper \\ \hline
\cellcolor[HTML]{DAE8FC}CAINNFlowCAC            & 98.38  & 97.11 & 98.40   & 99.13  & 95.80    & 99.62   & 96.71      & 96.30 & 94.81 & 96.89      & 96.62      & 94.56 & 98.64 & 97.33 & 98.69  \\ \hline
\cellcolor[HTML]{DAE8FC}CAINNFlowCA             & 98.03  & 96.71 & 98.51   & 98.98  & 95.98    & 99.54   & 96.53      & 95.52 & 94.41 & 96.88      & 96.31      & 94.25 & 98.06 & 97.26 & 98.19  \\ \hline
\cellcolor[HTML]{DAE8FC}CAINNFlowAC             & 98.07  & 95.48 & 98.50   & 98.95  & 95.22    & 99.60   & 95.17      & 94.43 & 93.44 & 96.30      & 96.12      & 94.73 & 98.07 & 97.25 & 98.35  \\ \hline
\cellcolor[HTML]{DAE8FC}CC                     & 98.06  & 96.82 & 98.65   & 99.11  & 95.78    & 99.58   & 95.21      & 97.10 & 94.56 & 96.43      & 96.34      & 95.42 & 98.57 & 96.15 & 98.57  \\ \hline
\end{tabular}}
\caption{\centering{Structure comparison of CAC, CA, AC and CC}\\
The feature extractor of the experiment was ResNet18, the epoch was $750$, the parameter was step2, and the learning rate LR was $5e-4$.
 }
\label{tab3}
\end{table*}
\begin{table*}[t]
\resizebox{\textwidth}{!}{
\begin{tabular}{|l|c|c|c|c|c|c|c|c|c|c|c|c|c|c|c|}
\hline
\rowcolor[HTML]{DAE8FC} 
                                          & bottle & cable & capsule & carpet & hazelnut & leather  & metal nut & pill  & tile  & toothbrush & transistor & wood  & grid  & screw & zipper \\ \hline
\cellcolor[HTML]{DAE8FC}CAINNFlowCAC/step5 & 97.88  & 96.65 & 97.99   & 98.83  & 96.15    & 99.58(8) & 93.48      & 95.92 & 93.50 & 94.85      & 96.53      & 93.92 & 98.85 & 95.93 & 98.31  \\ \hline
\cellcolor[HTML]{DAE8FC}CAINNFlowCAC/step4 & 97.99  & 96.76 & 98.17   & 98.99  & 96.02    & 99.58(5) & 93.86      & 95.10 & 93.85 & 93.34      & 96.85      & 93.80 & 98.27 & 95.99 & 98.15  \\ \hline
\cellcolor[HTML]{DAE8FC}CAINNFlowCAC/step3 & 98.01  & 96.91 & 98.12   & 99.11  & 95.20    & 99.65    & 95.40      & 95.77 & 94.89 & 96.69      & 96.74      & 94.45 & 98.57 & 96.70 & 98.46  \\ \hline
\cellcolor[HTML]{DAE8FC}CAINNFlowCAC/step2 & 98.38  & 97.11 & 98.40   & 99.13  & 95.80    & 99.62    & 96.71      & 96.30 & 94.81 & 96.89      & 96.62      & 94.56 & 98.64 & 97.33 & 98.69  \\ \hline
\cellcolor[HTML]{DAE8FC}CAINNFlowCAC/step1 & 98.28  & 96.33 & 98.64   & 99.02  & 96.74    & 99.55    & 97.15      & 95.73 & 94.22 & 97.82      & 95.80      & 95.03 & 98.12 & 96.93 & 98.51  \\ \hline
\end{tabular}}
\caption{  
Experimental results of CAC with different parametersResNet18 epoch was $750$, and the LR of learning was $5e-4$. }
\label{tab4}
\end{table*}
\subsection{contrast others}

\subsubsection{Different anomaly detection methods}


Table (\ref{tab1}): Anomaly detection and location performance of the MVTec AD dataset (image-level AUC, pixel-level AUC).
The MVTec AD dataset contains $5354$ high-resolution color images of different object and texture classes, the training set contains only normal samples without defects, and the test set adds abnormal samples. We compared folw with other anomaly detection methods that work well, such as: PatchSVDD \cite{yi2020patch}, SPADE \cite{reiss2021panda}, DifferNet \cite{rudolph2021same}, Cut Paste \cite{li2021cutpaste}, PaDiM \cite{defard2021padim}, PatchCore \cite{roth2021towards}, CFlow \cite{gudovskiy2022cflow}, CC \cite{yu2021fastflow}. The comparison results are shown in Table (\ref{tab1}). It can be found that in the case of Flow using fewer parameters, the image-level AUC and pixel-level AUC of most industrial products are higher than those of other anomaly detection methods.

\subsubsection{Different feature extractors}


As shown in Table (\ref{tab2}), we used four different feature extractors to detect and locate anomalies and filled the average into the table as the final result.
\begin{table}[]
\begin{tabular}{|
>{\columncolor[HTML]{DAE8FC}}l |c|c|c|c|}
\hline
AUROC   (Best) & \cellcolor[HTML]{DAE8FC}Wide-ResNet-50 & \cellcolor[HTML]{DAE8FC}ResNet18 & \cellcolor[HTML]{DAE8FC}DeiT & \cellcolor[HTML]{DAE8FC}CaiT \\ \hline
bottle         & 99.31                                  & 98.20                            & 96.64                        & 98.02                        \\ \hline
cable          & 98.44                                  & 96.87                            & 98.59                        & 98.63                        \\ \hline
capsule        & 98.89                                  & 98.41                            & 98.77                        & 98.92                        \\ \hline
carpet         & 99.15                                  & 98.76                            & 99.63                        & 99.31                        \\ \hline
grid           & 99.62                                  & 98.73                            & 98.34                        & 97.74                        \\ \hline
hazelnut       & 99.08                                  & 96.72                            & 99.11                        & 99.24                        \\ \hline
leather        & 99.62                                  & 99.62                            & 99.45                        & 99.64                        \\ \hline
metal nut      & 99.88                                  & 97.11                            & 99.92                        & 99.06                        \\ \hline
pill           & 97.31                                  & 96.26                            & 98.58                        & 98.79                        \\ \hline
screw          & 99.39                                  & 96.93                            & 99.48                        & 99.63                        \\ \hline
tile           & 97.42                                  & 94.81                            & 97.14                        & 97.48                        \\ \hline
toothbrush     & 99.11                                  & 97.76                            & 99.29                        & 99.53                        \\ \hline
transistor     & 97.51                                  & 97.53                            & 97.30                        & 97.22                        \\ \hline
wood           & 96.07                                  & 95.03                            & 95.99                        & 95.46                        \\ \hline
zipper         & 98.86                                  & 98.04                            & 97.62                        & 98.61                        \\ \hline
\textbf{Average}           & \textbf{98.64}                                  & \textbf{97.39}                            & \textbf{98.39}                        & \textbf{98.49}                        \\ \hline
\end{tabular}
\caption{}
\label{tab2}
\end{table}

The structure of CAINNFlow is a plug-in, and we can attach any feature extractor before it. To observe the performance of CAINNFlow when using different feature extractors, we designed a group of experiments, in which RESNet18, CaiT, wide-ResNet50-2 and Deit were used as our feature extractors, followed by CAINNFlow which used CAC structure to extract spatial information. Under the same parameter setting. It can be found that CAINNFlow using CaiT, Wide-ResNet50-2 and Deit feature extractor, the Pixels-level AUROC is generally higher than CAINNFlow using RESNet18 feature extractor. The experimental results show that the stronger the feature extractor before CAINNFlow, the better the output effect of CAINNFlow. The effect of the feature extractor and CAINNFlow combination structure is positively correlated with the strength of the feature extractor in the combination structure.

\subsection{Structure comparison of CAC, CA, AC and CC}

Our CAINNFlow has four structures for extracting spatial information: CA, AC, CAC, and CC. To compare the effects of CAINNFlow with four different structures, we conducted experiments on CAINNFlow using these four structures to extract spatial information, and the experimental results obtained are shown in Table (\ref{tab3}). In general, the CAC module adopted by CAINNFlow has a better overall effect than the CA module and AC module in $12$ evaluation indicators, and only the AC module is weaker than the CC module.

\begin{figure*}[t]
	\centering
	\includegraphics[width=.95\linewidth]{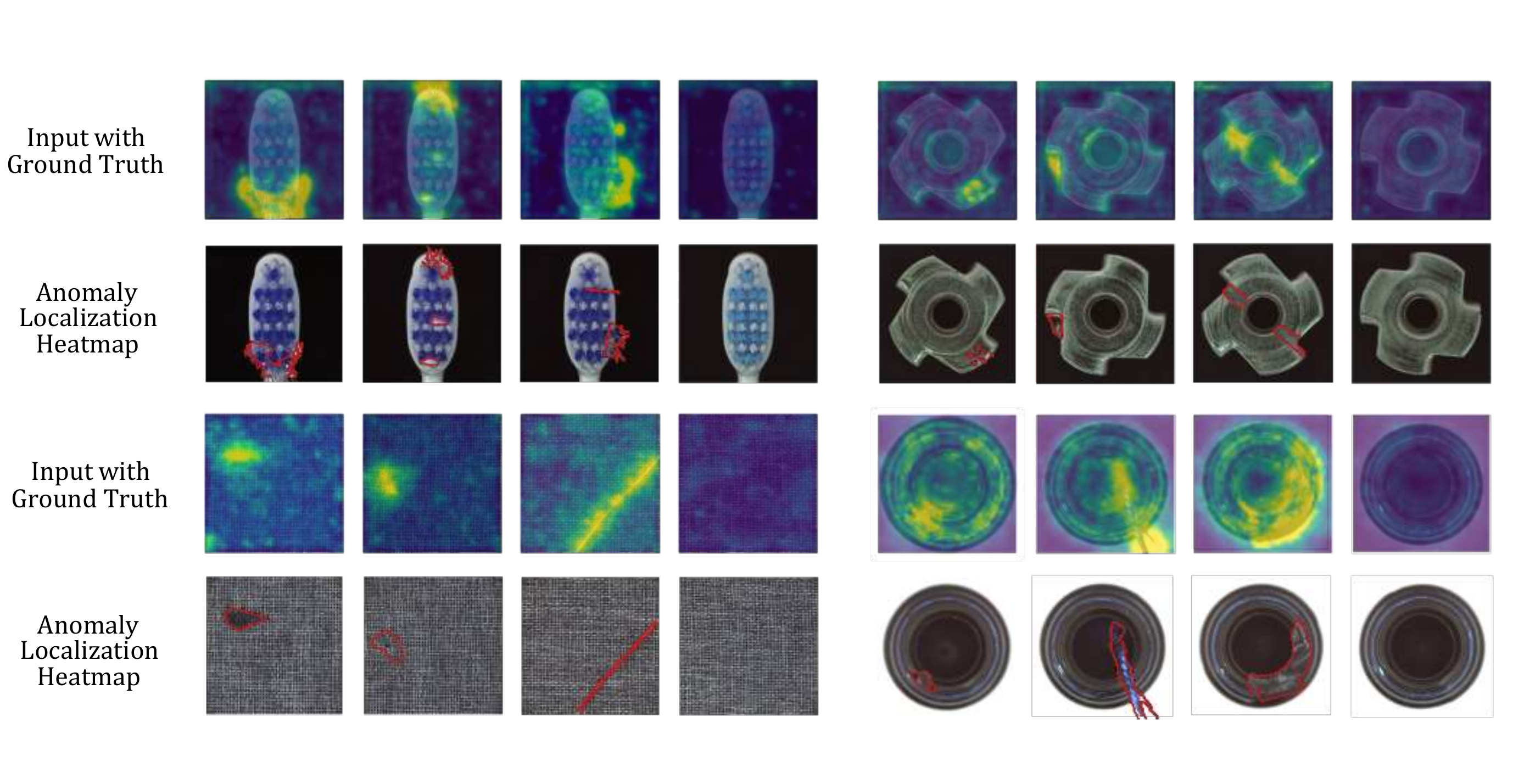}
	%\fbox{\rule[-.5cm]{4cm}{4cm} \rule[-.5cm]{4cm}{0cm}}
	\caption{Qualitative Results}
	\label{fig:fig11}
\end{figure*}

\vspace{-0.5em}
\subsection{CAC was compared with different steps}
We conducted experiments on the CAC module in $F$ under different parameters. In the same experiment, the feature extractor used RESNet18, the epoch was $750$, and the learning rate LR was $5e-4$. The experimental parameters were step1, Step2, step3, step4, and step5. The experimental results are shown in Table (\ref{tab4}):

It was observed that the evaluation index increased with Step and reached a better level when Step was equal to $2$, then decreased. In practice, Step can be set to $2$ to ensure performance and accuracy.

Although the performance of CAINNFlow with the large model as a feature extractor is better than that with the ResNet18 feature extractor, the experimental cost brought by the number of parameters of the large model is huge. To explore whether CAINNFlow can achieve better results when using small models as feature extractors, we conducted experiments. In the experiment, we used RESNet18 as the feature extractor and then modified the parameter Settings of CAINNFlow to find the global optimal solution. It is observed that under the condition of an invariable feature extractor, a better effect can be achieved by tuning CAINNFlow parameters. The experimental results are shown in Table (\ref{tab5}).

\begin{table}[]
\begin{tabular}{|
>{\columncolor[HTML]{DAE8FC}}c |c|
>{\columncolor[HTML]{DAE8FC}}c |c|}
\hline
Category   & \cellcolor[HTML]{DAE8FC}Best & Category                                      & \cellcolor[HTML]{DAE8FC}Best \\ \hline
bottle     & 98.38                        & pill                                          & 96.64                        \\ \hline
cable      & 97.47                        & tile                                          & 94.89                        \\ \hline
capsule    & 98.64                        & toothbrush                                    & 97.82                        \\ \hline
carpet     & 99.27                        & transistor                                    & 97.55                        \\ \hline
hazelnut   & 96.78                        & wood                                          & 95.03                        \\ \hline
leather    & 99.65                        & grid                                          & 98.85                        \\ \hline
metal nut & 97.15                        & screw                                         & 97.33                        \\ \hline
zipper     & 98.69                        & \multicolumn{1}{l|}{\cellcolor[HTML]{DAE8FC}} & \multicolumn{1}{l|}{}        \\ \hline
\end{tabular}
\caption{}
\label{tab5}
\end{table}

% \begin{table}[]
% \begin{tabular}{|
% >{\columncolor[HTML]{DAE8FC}}c |c|}
% \hline
%           & \cellcolor[HTML]{DAE8FC}BEST \\ \hline
% bottle     & 98.38                        \\ \hline
% cable      & 97.47                        \\ \hline
% capsule    & 98.64                        \\ \hline
% carpet     & 99.27                        \\ \hline
% hazelnut   & 96.78                        \\ \hline
% leather    & 99.65                        \\ \hline
% metal\_nut & 97.15                        \\ \hline
% pill       & 96.64                        \\ \hline
% tile       & 94.89                        \\ \hline
% toothbrush & 97.82                        \\ \hline
% transistor & 97.55                        \\ \hline
% wood       & 95.03                        \\ \hline
% grid       & 98.85                        \\ \hline
% screw      & 97.33                        \\ \hline
% zipper     & 98.69                        \\ \hline
% \end{tabular}
% \caption{}
% \label{tab5}
% \end{table}

% There are various bibliography styles available. You can select the
% style of your choice in the preamble of this document. These styles are
% Elsevier styles based on standard styles like Harvard and Vancouver.
% Please use Bib\TeX\ to generate your bibliography and include DOIs
% whenever available.

% Here are two sample references: 
% \cite{Fortunato2010}
% \cite{Fortunato2010,NewmanGirvan2004}
% \cite{Fortunato2010,Vehlowetal2013}
% \vspace{-0.5em}
\subsection{Feature Visualization and Generation}
The ability of bidirectional reversible probability distribution is one of the indispensable functions in our design of CAINNFlow. In the forward process, CAINNFlow can transform the feature map of the backbone network into the standard Gaussian distribution of high availability. In its reverse process, we also designed noise interference to prove that CAINNFlow has the visual characteristics of visualizing the reverse generation of specific probability sampling variables. In the figure(\ref{fig:fig10}) We have visualized this process.

As shown in figure(\ref{fig:fig10}) , as is shown in figure(\ref{fig:fig10}) In the forward process of class image, the probability graph after feature extraction and transformation has significant distribution characteristics. Moreover, in the experiment of adding noise points to the probability graph to verify the reversibility of CAINNFlow, obvious abnormal visual features were also obtained in the reverse results.
% \vspace{-0.5em}
\subsection{Qualitative Results}
Through experiments on the MVTecAD data set, we find the following figure(\ref{fig:fig11}) Visualizes the results. We designed a control experiment containing ground truth Mask and heatmap of abnormal location score with two kinds of reference images. The experimental results show that our CAINNFlow architecture is effective in anomaly recognition, whether based on normal images or abnormal images.

% \vspace{-1.5em}
\section{Conclusion}
CAINNFlow is a plug-in structure proposed in this study to identify and localize objects with anomalies, which can be used as a plug-in connected behind any computer vision model as a backbone feature extractor. In our research, in order to retain and effectively extract spatial structure information while performing distribution transformation, we constructed CAINNFlow based on INN by using a stacked CNN structure embedded with CBAM as a subnetwork.The results of extensive experiments on MVTec AD dataset show that CAINNFlow can be used as a feature extractor based on CNN and Transformer backbone network as a feature extractor, it achieves advanced level in terms of accuracy and inference efficiency.

\section*{Acknowledgements}
This research is supported by:  

$\bullet$ The National Natural Science Foundation of China (Grant No.41931183). The numerical calculation in this study were carried out on the SunRising-1 computing platform.

$\bullet$ The fund of the Beijing Municipal Education Commission, China, under grant number 22019821001.

$\bullet$ The fund of Climbing Program Foundation from Beijing Institute of Petrochemical Technology (Project No. BIPTAAI-2021007).

\bibliographystyle{model1-num-names}
% \bibliographystyle{cas-model2-names}

\bibliography{cas-refs}

\begin{thebibliography}{28}
\expandafter\ifx\csname natexlab\endcsname\relax\def\natexlab#1{#1}\fi
\providecommand{\url}[1]{\texttt{#1}}
\providecommand{\href}[2]{#2}
\providecommand{\path}[1]{#1}
\providecommand{\DOIprefix}{doi:}
\providecommand{\ArXivprefix}{arXiv:}
\providecommand{\URLprefix}{URL: }
\providecommand{\Pubmedprefix}{pmid:}
\providecommand{\doi}[1]{\href{http://dx.doi.org/#1}{\path{#1}}}
\providecommand{\Pubmed}[1]{\href{pmid:#1}{\path{#1}}}
\providecommand{\bibinfo}[2]{#2}
\ifx\xfnm\relax \def\xfnm[#1]{\unskip,\space#1}\fi
%Type = Article
\bibitem[{Roth et~al.(2021)Roth, Pemula, Zepeda, Sch{\"o}lkopf, Brox, and
  Gehler}]{roth2021towards}
\bibinfo{author}{K.~Roth}, \bibinfo{author}{L.~Pemula},
  \bibinfo{author}{J.~Zepeda}, \bibinfo{author}{B.~Sch{\"o}lkopf},
  \bibinfo{author}{T.~Brox}, \bibinfo{author}{P.~Gehler},
\newblock \bibinfo{title}{Towards total recall in industrial anomaly
  detection},
\newblock \bibinfo{journal}{arXiv preprint arXiv:2106.08265}
  (\bibinfo{year}{2021}).
%Type = Article
\bibitem[{Bergman and Hoshen(2020)}]{bergman2020classification}
\bibinfo{author}{L.~Bergman}, \bibinfo{author}{Y.~Hoshen},
\newblock \bibinfo{title}{Classification-based anomaly detection for general
  data},
\newblock \bibinfo{journal}{arXiv preprint arXiv:2005.02359}
  (\bibinfo{year}{2020}).
%Type = Inproceedings
\bibitem[{Wang et~al.(2021)Wang, Wu, Cui, and Shen}]{wang2021glancing}
\bibinfo{author}{S.~Wang}, \bibinfo{author}{L.~Wu}, \bibinfo{author}{L.~Cui},
  \bibinfo{author}{Y.~Shen},
\newblock \bibinfo{title}{Glancing at the patch: Anomaly localization with
  global and local feature comparison},
\newblock in: \bibinfo{booktitle}{Proceedings of the IEEE/CVF Conference on
  Computer Vision and Pattern Recognition}, \bibinfo{year}{2021}, pp.
  \bibinfo{pages}{254--263}.
%Type = Inproceedings
\bibitem[{Li et~al.(2021)Li, Sohn, Yoon, and Pfister}]{li2021cutpaste}
\bibinfo{author}{C.-L. Li}, \bibinfo{author}{K.~Sohn},
  \bibinfo{author}{J.~Yoon}, \bibinfo{author}{T.~Pfister},
\newblock \bibinfo{title}{Cutpaste: Self-supervised learning for anomaly
  detection and localization},
\newblock in: \bibinfo{booktitle}{Proceedings of the IEEE/CVF Conference on
  Computer Vision and Pattern Recognition}, \bibinfo{year}{2021}, pp.
  \bibinfo{pages}{9664--9674}.
%Type = Article
\bibitem[{Dosovitskiy et~al.(2020)Dosovitskiy, Beyer, Kolesnikov, Weissenborn,
  Zhai, Unterthiner, Dehghani, Minderer, Heigold, Gelly
  et~al.}]{dosovitskiy2020image}
\bibinfo{author}{A.~Dosovitskiy}, \bibinfo{author}{L.~Beyer},
  \bibinfo{author}{A.~Kolesnikov}, \bibinfo{author}{D.~Weissenborn},
  \bibinfo{author}{X.~Zhai}, \bibinfo{author}{T.~Unterthiner},
  \bibinfo{author}{M.~Dehghani}, \bibinfo{author}{M.~Minderer},
  \bibinfo{author}{G.~Heigold}, \bibinfo{author}{S.~Gelly}, et~al.,
\newblock \bibinfo{title}{An image is worth 16x16 words: Transformers for image
  recognition at scale},
\newblock \bibinfo{journal}{arXiv preprint arXiv:2010.11929}
  (\bibinfo{year}{2020}).
%Type = Inproceedings
\bibitem[{Liu et~al.(2021)Liu, Lin, Cao, Hu, Wei, Zhang, Lin, and
  Guo}]{liu2021Swin}
\bibinfo{author}{Z.~Liu}, \bibinfo{author}{Y.~Lin}, \bibinfo{author}{Y.~Cao},
  \bibinfo{author}{H.~Hu}, \bibinfo{author}{Y.~Wei},
  \bibinfo{author}{Z.~Zhang}, \bibinfo{author}{S.~Lin},
  \bibinfo{author}{B.~Guo},
\newblock \bibinfo{title}{Swin transformer: Hierarchical vision transformer
  using shifted windows},
\newblock in: \bibinfo{booktitle}{Proceedings of the IEEE/CVF International
  Conference on Computer Vision (ICCV)}, \bibinfo{year}{2021}.
%Type = Article
\bibitem[{Cunningham et~al.(2020)Cunningham, Zabounidis, Agrawal, Fiterau, and
  Sheldon}]{cunningham2020normalizing}
\bibinfo{author}{E.~Cunningham}, \bibinfo{author}{R.~Zabounidis},
  \bibinfo{author}{A.~Agrawal}, \bibinfo{author}{I.~Fiterau},
  \bibinfo{author}{D.~Sheldon},
\newblock \bibinfo{title}{Normalizing flows across dimensions},
\newblock \bibinfo{journal}{arXiv preprint arXiv:2006.13070}
  (\bibinfo{year}{2020}).
%Type = Inproceedings
\bibitem[{Woo et~al.(2018)Woo, Park, Lee, and Kweon}]{woo2018cbam}
\bibinfo{author}{S.~Woo}, \bibinfo{author}{J.~Park}, \bibinfo{author}{J.-Y.
  Lee}, \bibinfo{author}{I.~S. Kweon},
\newblock \bibinfo{title}{Cbam: Convolutional block attention module},
\newblock in: \bibinfo{booktitle}{Proceedings of the European conference on
  computer vision (ECCV)}, \bibinfo{year}{2018}, pp. \bibinfo{pages}{3--19}.
%Type = Inproceedings
\bibitem[{Bergmann et~al.(2019)Bergmann, Fauser, Sattlegger, and
  Steger}]{bergmann2019mvtec}
\bibinfo{author}{P.~Bergmann}, \bibinfo{author}{M.~Fauser},
  \bibinfo{author}{D.~Sattlegger}, \bibinfo{author}{C.~Steger},
\newblock \bibinfo{title}{Mvtec ad--a comprehensive real-world dataset for
  unsupervised anomaly detection},
\newblock in: \bibinfo{booktitle}{Proceedings of the IEEE/CVF conference on
  computer vision and pattern recognition}, \bibinfo{year}{2019}, pp.
  \bibinfo{pages}{9592--9600}.
%Type = Article
\bibitem[{Feng et~al.(2017)Feng, Yuan, and Lu}]{feng2017learning}
\bibinfo{author}{Y.~Feng}, \bibinfo{author}{Y.~Yuan}, \bibinfo{author}{X.~Lu},
\newblock \bibinfo{title}{Learning deep event models for crowd anomaly
  detection},
\newblock \bibinfo{journal}{Neurocomputing} \bibinfo{volume}{219}
  (\bibinfo{year}{2017}) \bibinfo{pages}{548--556}.
%Type = Article
\bibitem[{Wang et~al.(2018)Wang, Zhu, Yin, and Porikli}]{wang2018video}
\bibinfo{author}{S.~Wang}, \bibinfo{author}{E.~Zhu}, \bibinfo{author}{J.~Yin},
  \bibinfo{author}{F.~Porikli},
\newblock \bibinfo{title}{Video anomaly detection and localization by local
  motion based joint video representation and ocelm},
\newblock \bibinfo{journal}{Neurocomputing} \bibinfo{volume}{277}
  (\bibinfo{year}{2018}) \bibinfo{pages}{161--175}.
%Type = Article
\bibitem[{Li and Chang(2019)}]{li2019video}
\bibinfo{author}{N.~Li}, \bibinfo{author}{F.~Chang},
\newblock \bibinfo{title}{Video anomaly detection and localization via
  multivariate gaussian fully convolution adversarial autoencoder},
\newblock \bibinfo{journal}{Neurocomputing} \bibinfo{volume}{369}
  (\bibinfo{year}{2019}) \bibinfo{pages}{92--105}.
%Type = Article
\bibitem[{Canizo et~al.(2019)Canizo, Triguero, Conde, and
  Onieva}]{canizo2019multi}
\bibinfo{author}{M.~Canizo}, \bibinfo{author}{I.~Triguero},
  \bibinfo{author}{A.~Conde}, \bibinfo{author}{E.~Onieva},
\newblock \bibinfo{title}{Multi-head cnn--rnn for multi-time series anomaly
  detection: An industrial case study},
\newblock \bibinfo{journal}{Neurocomputing} \bibinfo{volume}{363}
  (\bibinfo{year}{2019}) \bibinfo{pages}{246--260}.
%Type = Article
\bibitem[{Zhang et~al.(2020)Zhang, Ma, Yu, Huang, Howell, and
  Stevens}]{zhang2020scene}
\bibinfo{author}{X.~Zhang}, \bibinfo{author}{D.~Ma}, \bibinfo{author}{H.~Yu},
  \bibinfo{author}{Y.~Huang}, \bibinfo{author}{P.~Howell},
  \bibinfo{author}{B.~Stevens},
\newblock \bibinfo{title}{Scene perception guided crowd anomaly detection},
\newblock \bibinfo{journal}{Neurocomputing} \bibinfo{volume}{414}
  (\bibinfo{year}{2020}) \bibinfo{pages}{291--302}.
%Type = Article
\bibitem[{Fahad and Tahir(2021)}]{fahad2021activity}
\bibinfo{author}{L.~G. Fahad}, \bibinfo{author}{S.~F. Tahir},
\newblock \bibinfo{title}{Activity recognition and anomaly detection in smart
  homes},
\newblock \bibinfo{journal}{Neurocomputing} \bibinfo{volume}{423}
  (\bibinfo{year}{2021}) \bibinfo{pages}{362--372}.
%Type = Article
\bibitem[{Yu et~al.(2021)Yu, Zheng, Wang, Li, Wu, Zhao, and
  Wu}]{yu2021fastflow}
\bibinfo{author}{J.~Yu}, \bibinfo{author}{Y.~Zheng}, \bibinfo{author}{X.~Wang},
  \bibinfo{author}{W.~Li}, \bibinfo{author}{Y.~Wu}, \bibinfo{author}{R.~Zhao},
  \bibinfo{author}{L.~Wu},
\newblock \bibinfo{title}{Fastflow: Unsupervised anomaly detection and
  localization via 2d normalizing flows},
\newblock \bibinfo{journal}{arXiv preprint arXiv:2111.07677}
  (\bibinfo{year}{2021}).
%Type = Article
\bibitem[{Li et~al.(2016)Li, Jiao, Han, and Weissman}]{li2016demystifying}
\bibinfo{author}{S.~Li}, \bibinfo{author}{J.~Jiao}, \bibinfo{author}{Y.~Han},
  \bibinfo{author}{T.~Weissman},
\newblock \bibinfo{title}{Demystifying resnet},
\newblock \bibinfo{journal}{arXiv preprint arXiv:1611.01186}
  (\bibinfo{year}{2016}).
%Type = Article
\bibitem[{Simonyan and Zisserman(2014)}]{simonyan2014very}
\bibinfo{author}{K.~Simonyan}, \bibinfo{author}{A.~Zisserman},
\newblock \bibinfo{title}{Very deep convolutional networks for large-scale
  image recognition},
\newblock \bibinfo{journal}{arXiv preprint arXiv:1409.1556}
  (\bibinfo{year}{2014}).
%Type = Inproceedings
\bibitem[{Touvron et~al.(2021{\natexlab{a}})Touvron, Cord, Douze, Massa,
  Sablayrolles, and J{\'e}gou}]{touvron2021training}
\bibinfo{author}{H.~Touvron}, \bibinfo{author}{M.~Cord},
  \bibinfo{author}{M.~Douze}, \bibinfo{author}{F.~Massa},
  \bibinfo{author}{A.~Sablayrolles}, \bibinfo{author}{H.~J{\'e}gou},
\newblock \bibinfo{title}{Training data-efficient image transformers \&
  distillation through attention},
\newblock in: \bibinfo{booktitle}{International Conference on Machine
  Learning}, \bibinfo{organization}{PMLR}, \bibinfo{year}{2021}{\natexlab{a}},
  pp. \bibinfo{pages}{10347--10357}.
%Type = Inproceedings
\bibitem[{Touvron et~al.(2021{\natexlab{b}})Touvron, Cord, Sablayrolles,
  Synnaeve, and J{\'e}gou}]{touvron2021going}
\bibinfo{author}{H.~Touvron}, \bibinfo{author}{M.~Cord},
  \bibinfo{author}{A.~Sablayrolles}, \bibinfo{author}{G.~Synnaeve},
  \bibinfo{author}{H.~J{\'e}gou},
\newblock \bibinfo{title}{Going deeper with image transformers},
\newblock in: \bibinfo{booktitle}{Proceedings of the IEEE/CVF International
  Conference on Computer Vision}, \bibinfo{year}{2021}{\natexlab{b}}, pp.
  \bibinfo{pages}{32--42}.
%Type = Article
\bibitem[{Liu et~al.(2022)Liu, Mao, Wu, Feichtenhofer, Darrell, and
  Xie}]{liu2022convnet}
\bibinfo{author}{Z.~Liu}, \bibinfo{author}{H.~Mao}, \bibinfo{author}{C.-Y. Wu},
  \bibinfo{author}{C.~Feichtenhofer}, \bibinfo{author}{T.~Darrell},
  \bibinfo{author}{S.~Xie},
\newblock \bibinfo{title}{A convnet for the 2020s},
\newblock \bibinfo{journal}{arXiv preprint arXiv:2201.03545}
  (\bibinfo{year}{2022}).
%Type = Article
\bibitem[{Agnelli et~al.(2010)Agnelli, Cadeiras, Tabak, Turner, and
  Vanden-Eijnden}]{agnelli2010clustering}
\bibinfo{author}{J.~Agnelli}, \bibinfo{author}{M.~Cadeiras},
  \bibinfo{author}{E.~G. Tabak}, \bibinfo{author}{C.~V. Turner},
  \bibinfo{author}{E.~Vanden-Eijnden},
\newblock \bibinfo{title}{Clustering and classification through normalizing
  flows in feature space},
\newblock \bibinfo{journal}{Multiscale Modeling \& Simulation}
  \bibinfo{volume}{8} (\bibinfo{year}{2010}) \bibinfo{pages}{1784--1802}.
%Type = Article
\bibitem[{Kobyzev et~al.(2020)Kobyzev, Prince, and
  Brubaker}]{kobyzev2020normalizing}
\bibinfo{author}{I.~Kobyzev}, \bibinfo{author}{S.~J. Prince},
  \bibinfo{author}{M.~A. Brubaker},
\newblock \bibinfo{title}{Normalizing flows: An introduction and review of
  current methods},
\newblock \bibinfo{journal}{IEEE transactions on pattern analysis and machine
  intelligence} \bibinfo{volume}{43} (\bibinfo{year}{2020})
  \bibinfo{pages}{3964--3979}.
%Type = Inproceedings
\bibitem[{Yi and Yoon(2020)}]{yi2020patch}
\bibinfo{author}{J.~Yi}, \bibinfo{author}{S.~Yoon},
\newblock \bibinfo{title}{Patch svdd: Patch-level svdd for anomaly detection
  and segmentation},
\newblock in: \bibinfo{booktitle}{Proceedings of the Asian Conference on
  Computer Vision}, \bibinfo{year}{2020}.
%Type = Inproceedings
\bibitem[{Reiss et~al.(2021)Reiss, Cohen, Bergman, and Hoshen}]{reiss2021panda}
\bibinfo{author}{T.~Reiss}, \bibinfo{author}{N.~Cohen},
  \bibinfo{author}{L.~Bergman}, \bibinfo{author}{Y.~Hoshen},
\newblock \bibinfo{title}{Panda: Adapting pretrained features for anomaly
  detection and segmentation},
\newblock in: \bibinfo{booktitle}{Proceedings of the IEEE/CVF Conference on
  Computer Vision and Pattern Recognition}, \bibinfo{year}{2021}, pp.
  \bibinfo{pages}{2806--2814}.
%Type = Inproceedings
\bibitem[{Rudolph et~al.(2021)Rudolph, Wandt, and Rosenhahn}]{rudolph2021same}
\bibinfo{author}{M.~Rudolph}, \bibinfo{author}{B.~Wandt},
  \bibinfo{author}{B.~Rosenhahn},
\newblock \bibinfo{title}{Same same but differnet: Semi-supervised defect
  detection with normalizing flows},
\newblock in: \bibinfo{booktitle}{Proceedings of the IEEE/CVF Winter Conference
  on Applications of Computer Vision}, \bibinfo{year}{2021}, pp.
  \bibinfo{pages}{1907--1916}.
%Type = Inproceedings
\bibitem[{Defard et~al.(2021)Defard, Setkov, Loesch, and
  Audigier}]{defard2021padim}
\bibinfo{author}{T.~Defard}, \bibinfo{author}{A.~Setkov},
  \bibinfo{author}{A.~Loesch}, \bibinfo{author}{R.~Audigier},
\newblock \bibinfo{title}{Padim: a patch distribution modeling framework for
  anomaly detection and localization},
\newblock in: \bibinfo{booktitle}{International Conference on Pattern
  Recognition}, \bibinfo{organization}{Springer}, \bibinfo{year}{2021}, pp.
  \bibinfo{pages}{475--489}.
%Type = Inproceedings
\bibitem[{Gudovskiy et~al.(2022)Gudovskiy, Ishizaka, and
  Kozuka}]{gudovskiy2022cflow}
\bibinfo{author}{D.~Gudovskiy}, \bibinfo{author}{S.~Ishizaka},
  \bibinfo{author}{K.~Kozuka},
\newblock \bibinfo{title}{Cflow-ad: Real-time unsupervised anomaly detection
  with localization via conditional normalizing flows},
\newblock in: \bibinfo{booktitle}{Proceedings of the IEEE/CVF Winter Conference
  on Applications of Computer Vision}, \bibinfo{year}{2022}, pp.
  \bibinfo{pages}{98--107}.

\end{thebibliography}

% \vskip3pt

% Author biography without author photo.
% Author biography. Author biography. Author biography.
% Author biography. Author biography. Author biography.
% Author biography. Author biography. Author biography.
% Author biography. Author biography. Author biography.
% Author biography. Author biography. Author biography.
% Author biography. Author biography. Author biography.
% Author biography. Author biography. Author biography.
% Author biography. Author biography. Author biography.
% Author biography. Author biography. Author biography.
%强制换页代码
\newpage
\newpage
\bio{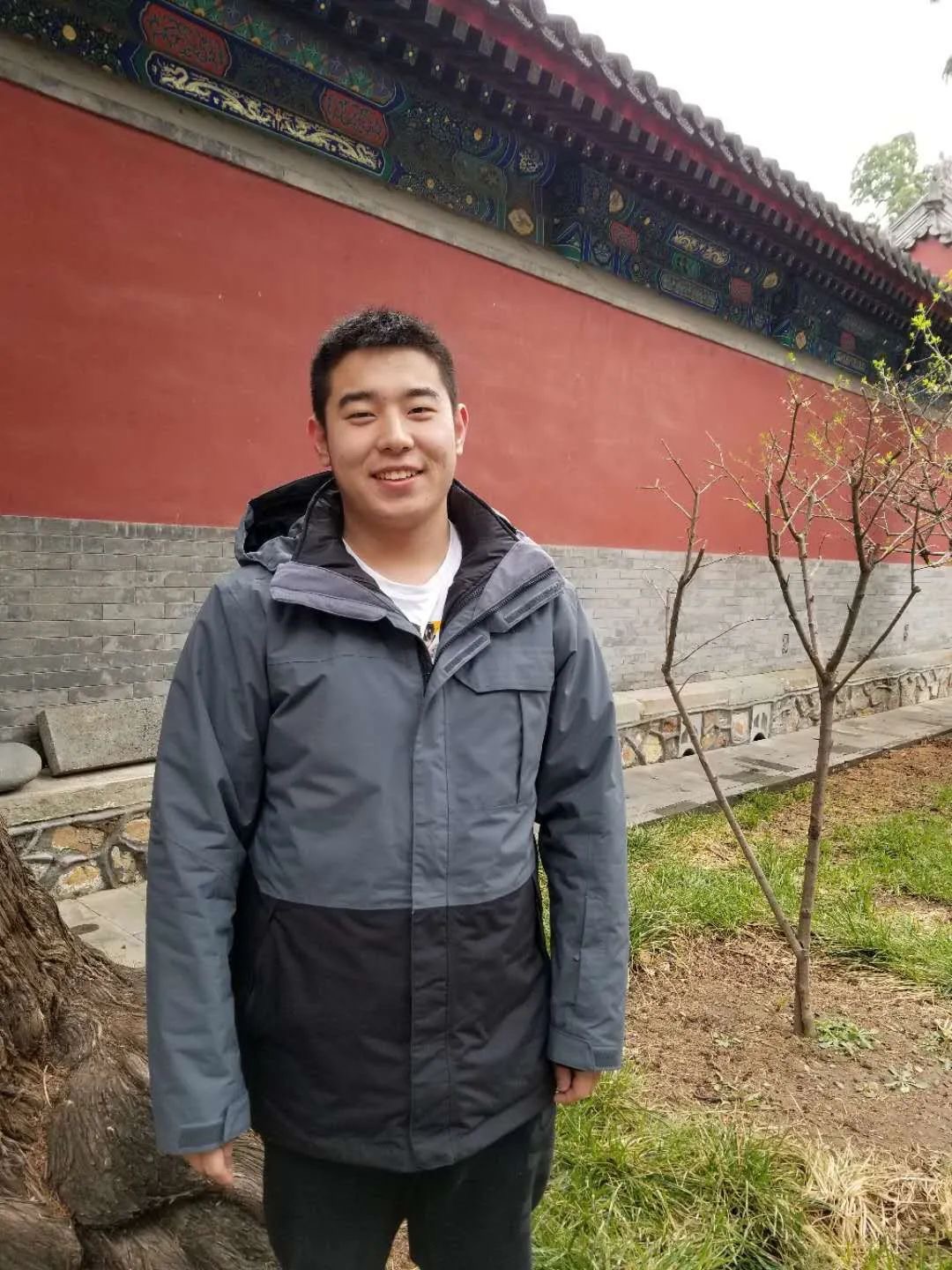}
Yan Ruiqing, born in 2000, is an undergraduate student in Beijing Institute of Petrochemical Technology. He is now working as an intern in computer Network Information Center of Chinese Academy of Sciences, engaged in algorithm research. His research interests include computer vision, natural language processing, spatio-temporal modeling algorithms, and cross-modal modeling algorithms.
\endbio
\vspace{1.5em}
\bio{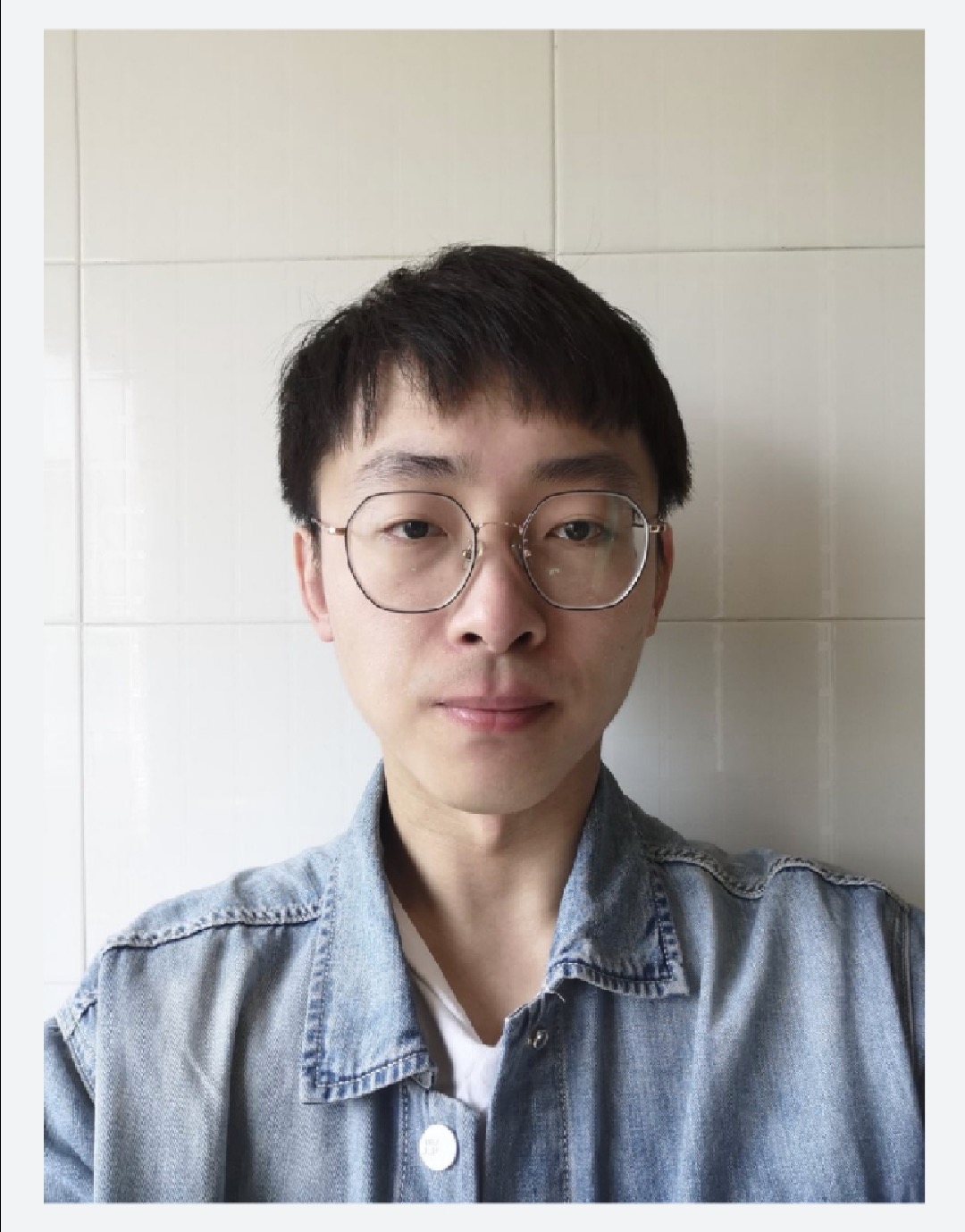}
Linghan Zheng received the Master's degree in Tongji University, China in 2017. He now works for Ant Group and works on natural language understanding. His research interests is natural language retrival and natural language representation problem. At the same time, he also has a strong interest in multimodality content understanding.
\endbio
%改变行距代码
\vspace{3em}
\bio{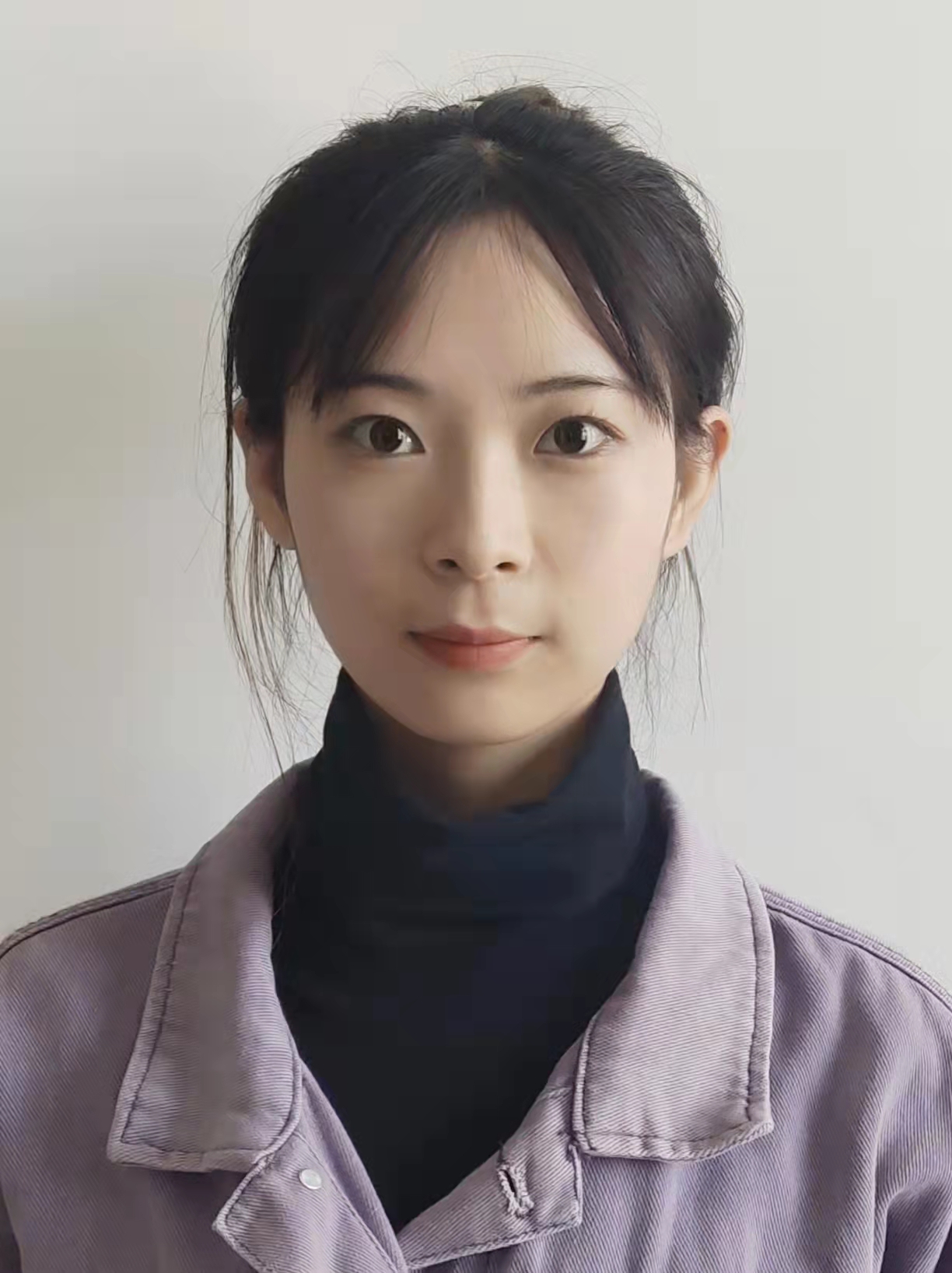}
Jinfeng Li received a bachelor's degree in software engineering from Taiyuan University of Science and Technology in 2021 and is currently pursuing a master's degree in the School of Information Engineering, Beijing Institute of Petrochemical Technology. Her research interests are in computer vision.
\endbio
\vspace{2.5em}
\bio{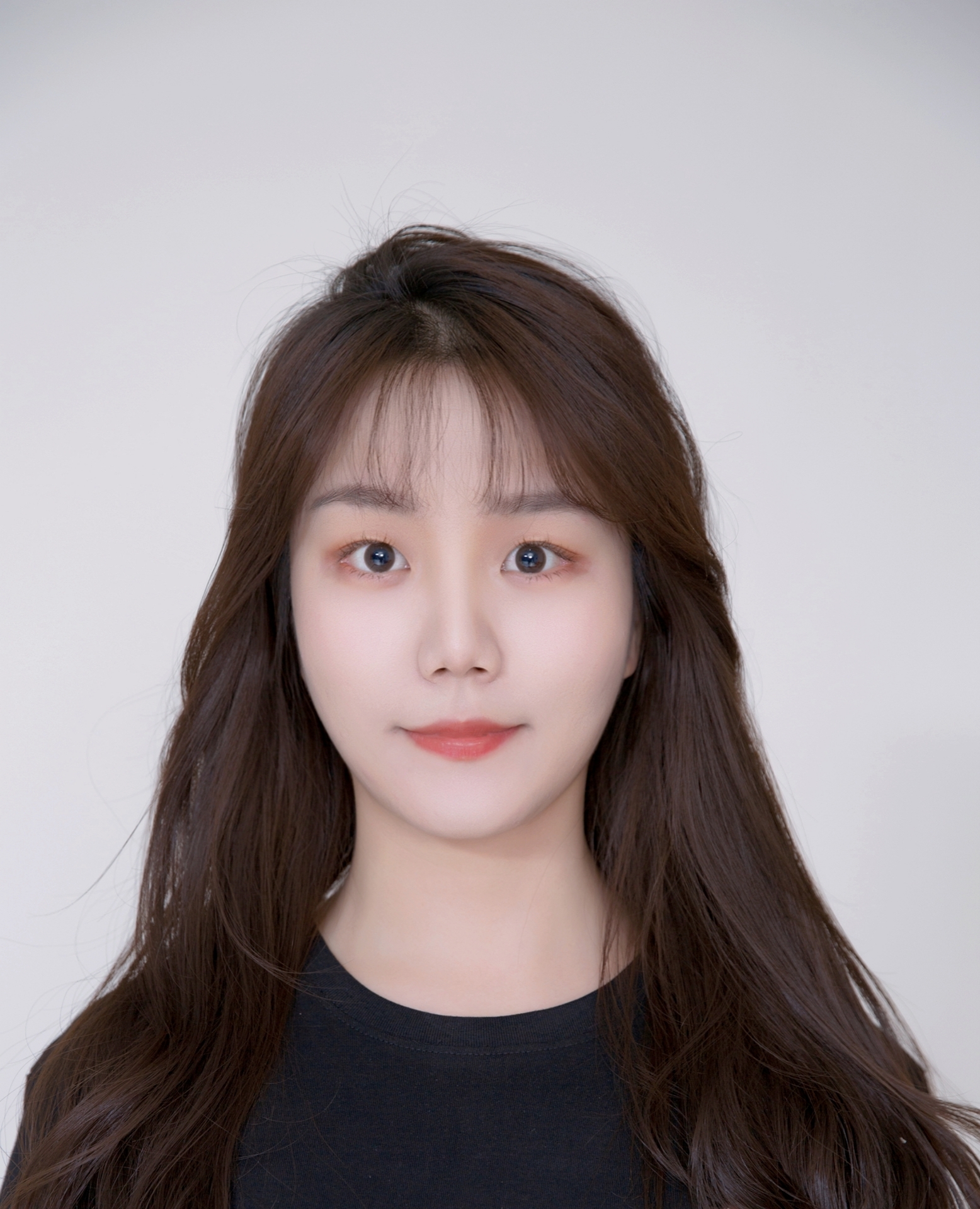}
Mengyuan Huang , graduated from Heilongjiang University with a bachelor's degree in network engineering in 2019. From 2019 to 2020, she worked as a software engineer in Shanghai BYD Co., Ltd. Since 2021, she has been studying at the Artificial Intelligence Research Institute of Beijing Institute of Petrochemical Technology as a graduate student. Her current research interests include smart healthcare.
\endbio
\vspace{-0.7em}
\bio{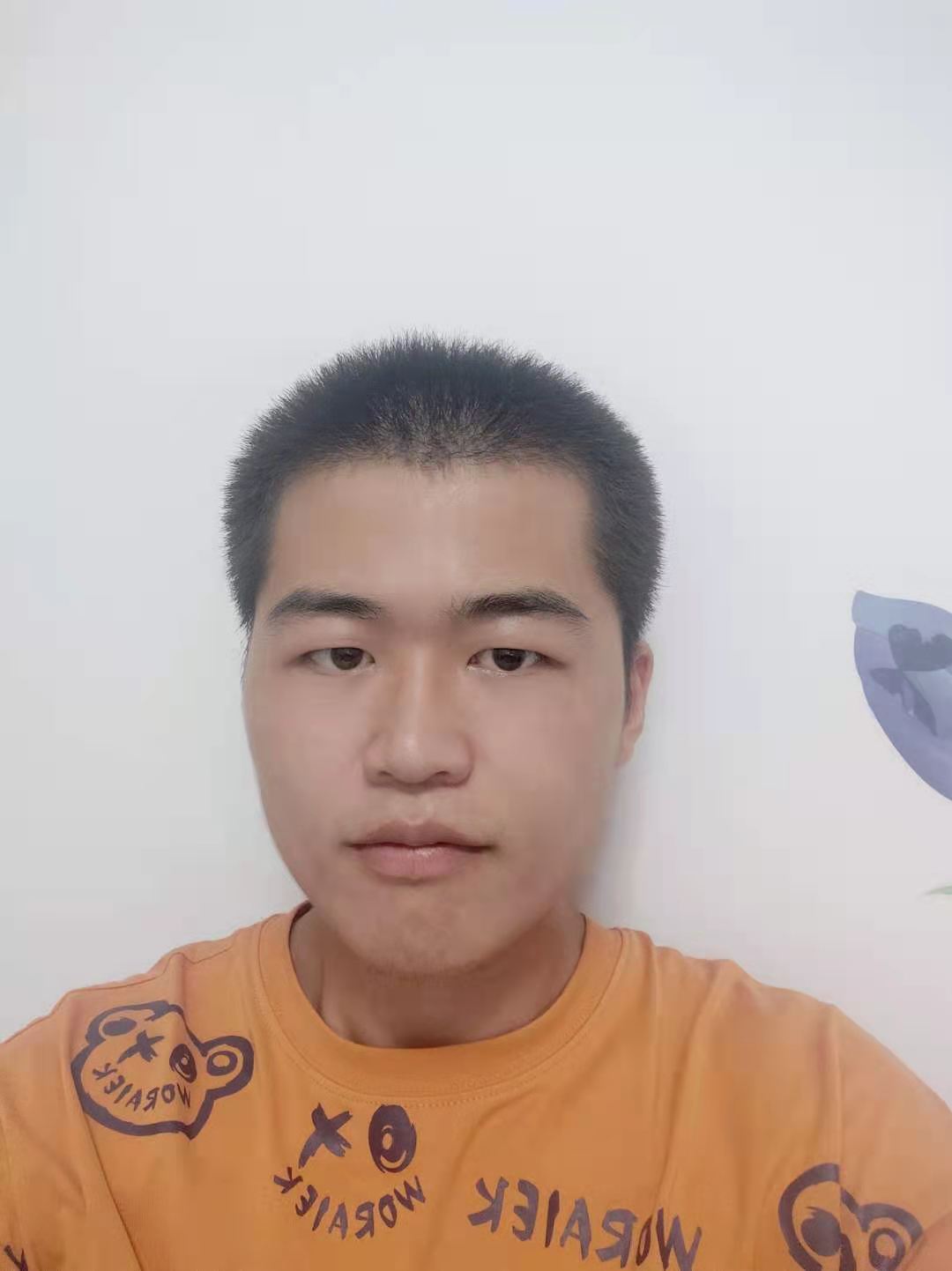}
Liu Wu received his B.Eng. degree in high polymer material from Hubei University in 2019.From 2019 to 2021,Engaged in python development. Be familiared with Django framework and VUE. Organized and built e-commerce websites, novel websites and crawler projects during career.His research interests include computer vision.
\endbio
\vspace{2.5em}
\bio{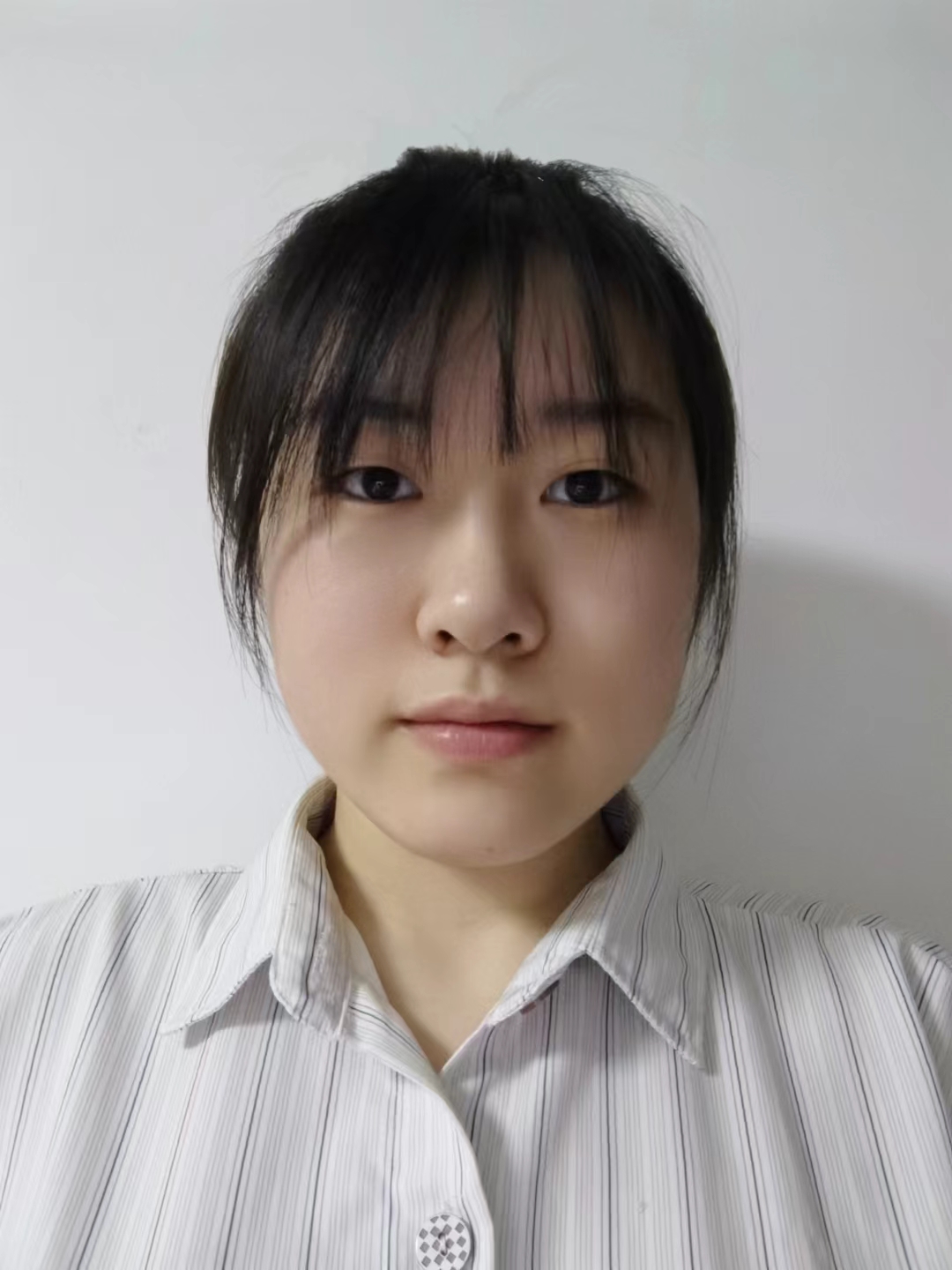}
Dongyu Hu, from 2020 to now, is studying for a bachelor's degree in computer science and technology at the School of Information Engineering, Beijing Institute of Petrochemical Technology. During school, she has participated in a number of artificial intelligence projects. Her current research interests include artificial intelligence, computer vision and intelligent medical care.
\endbio
\vspace{1.5em}
\newpage
\bio{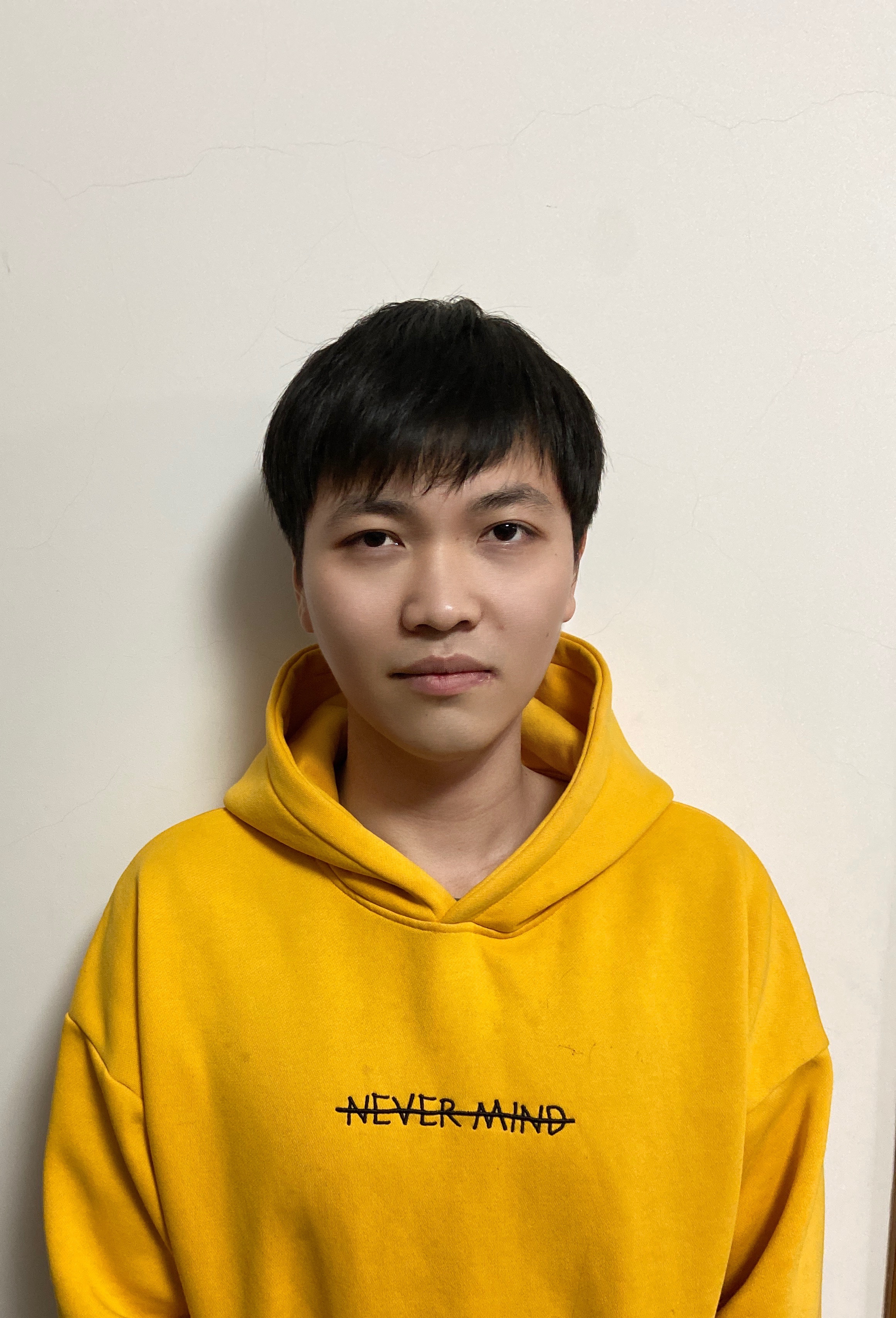}
Zhang Fan received the bachelor's degree in Jiyang College of Zhejiang agriculture and Forestry University ,China in 2020.From March 2020 to May 2020,he worked in software testing.From 2021 to now, he is a graduate student at the  Department of Information Engineering of Beijing Institute of Petrochemical Technology. His current research interests include natural language processing and image processing.
\endbio
\vspace{1.5em}
\bio{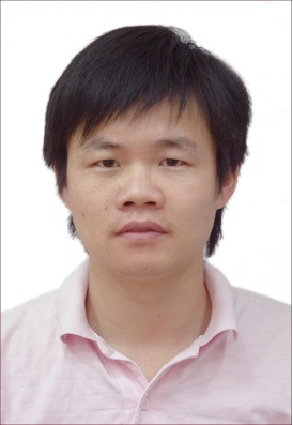}
Jinrong Jiang received the Ph.D. degree in enginering frim Chinese Academy of Sciences in 2007. He is a professor at Computer Network Information Center,Chinese Academy of Sciences, Beijing, China. His research interests include high performance computing, computational geosciences and AI application.
\endbio
\vspace{2.5em}
\bio{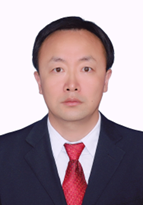}
Qianjin Guo received the Ph.D. degree in Mechanical and Electronic Engineering from Shenyang Institute of Automation (SIA), Chinese Academy of Sciences, in 2008. From 2009 to 2021, he was a Associate Researcher with the Institute of Chemistry,Chinese Academy of Sciences.Since 2021, he is working as a Full Professor at the School of Academy of Artificial Intelligence, Beijing Institute of Petrochemical Technology, Beijing , China. His current research interests include artificial intelligent, soft computing, deep learning, intelligent and hybrid control systems, signal processing, biomedical imaging and chemical imaging in innovation. 
\endbio
\vspace{-1.5em}
\bio{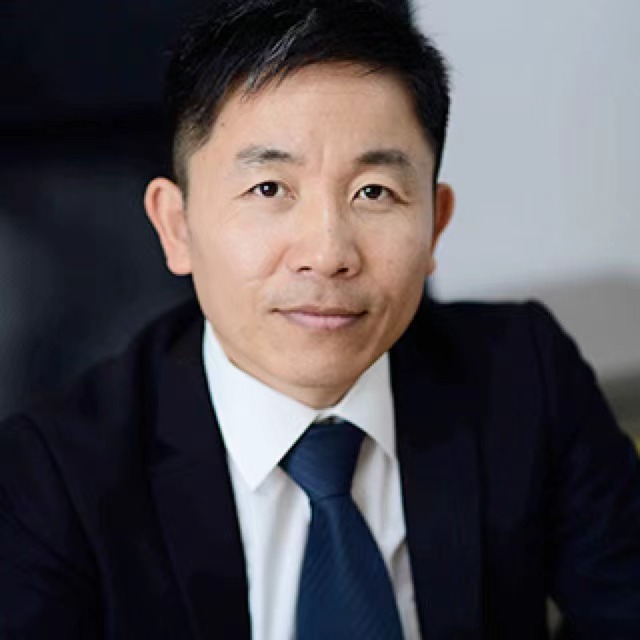}
Dr. Liu Qiang, professor, male, artificial intelligence and integrated circuit expert, CISP. September 1996 - July 2000, Majored in Computer Software and Theory, Department of Computer Science and Technology, Peking University, bachelor's degree. September 2000 -- July 2005, Computer System Structure, Department of Computer Science and Technology, Peking University, PH. D. degree.
He has worked in IBM China Research Institute and China Electronics Information Industry Group and held senior management positions. Participated in the establishment of four high-tech enterprises, two research and development institutions, and one investment institution. 
% Main research projects include: 
% (1) Jbcore16, Unity805, and Unity863 Zhongzhi series microprocessor and system chip. The latest technical expertise to participate in the development of general CPU in China, one of the main technical leaders of "China Chip"; 
% (2) chip of public-key cryptosystem and security system; 
% (3) RESEARCH on IBM cloud computing and cloud security system. Responsible for the development of multimedia processor and network processor, the key technology research results have been at the international leading level for many years, and has been widely used in cloud computing information system; 
% (4) Digital media content analysis and extraction system; 
% (5) Supercomputer network content processing system. 
\endbio

% \vspace{2em}

% \bio{figs/pic1}
% Author biography with author photo.
% Author biography. Author biography. Author biography.
% Author biography. Author biography. Author biography.
% Author biography. Author biography. Author biography.
% Author biography. Author biography. Author biography.
\end{sloppypar}
\end{document}